\documentclass[10pt,twocolumn,letterpaper]{article}

\usepackage{cvpr}
\usepackage{times}
\usepackage{epsfig}
\usepackage{graphicx}
\usepackage{amsmath}
\usepackage{amssymb}
\usepackage{color}
\usepackage{multirow}
\usepackage{caption} 
\usepackage[breaklinks=true,bookmarks=false]{hyperref}
\hypersetup{
    colorlinks=true,     
    urlcolor=magenta
}
\usepackage{siunitx}
\usepackage{textcomp}
\usepackage{enumitem}
\usepackage{makecell}
\usepackage[section]{placeins}
\usepackage[title,toc,titletoc,page]{appendix}
\usepackage{threeparttable}
\usepackage{array}
\newcolumntype{?}{!{\vrule width 1pt}}
\graphicspath{ {figures/} }

\cvprfinalcopy % *** Uncomment this line for the final submission

\def\cvprPaperID{5302} % *** Enter the CVPR Paper ID here

\begin{document}

%%%%%%%%% TITLE
\title{KFNet: Learning Temporal Camera Relocalization using Kalman Filtering}

\author{
Lei Zhou$^{1}$ \hspace{1cm}
Zixin Luo$^{1}$ \hspace{1cm}
Tianwei Shen$^{1}$ \hspace{1cm}
Jiahui Zhang$^{2}$ \\
Mingmin Zhen$^{1}$ \hspace{1cm}
Yao Yao$^{1}$ \hspace{1cm}
Tian Fang$^{3}$ \hspace{1cm}
Long Quan$^{1}$ \\
\normalsize $^1$Hong Kong University of Science and Technology  \hspace{0.3cm} 
\normalsize $^2$Tsinghua University \hspace{0.3cm}
\normalsize $^3$Everest Innovation Technology
\\
\tt\small $^1$\{lzhouai,zluoag,tshenaa,mzhen,yyaoag,quan\}@cse.ust.hk \hspace{0.3cm} \\
\tt\small $^2$jiahui-z15@mails.tsinghua.edu.cn\hspace{0.7cm}
$^3$fangtian@altizure.com
}

\maketitle
\begin{abstract}
Temporal camera relocalization estimates the pose with respect to each video frame in sequence, as opposed to one-shot relocalization which focuses on a still image.
Even though the time dependency has been taken into account, current temporal relocalization methods still generally underperform the state-of-the-art one-shot approaches in terms of accuracy.
In this work, we improve the temporal relocalization method by using a network architecture that
incorporates Kalman filtering (KFNet) for online camera relocalization. 
In particular, KFNet extends the scene coordinate regression problem to the time domain in order to recursively establish 2D and 3D correspondences for the pose determination.
The network architecture design and the loss formulation are based on Kalman filtering in the context of Bayesian learning. 
Extensive experiments on multiple relocalization benchmarks demonstrate the high accuracy of KFNet at the top of both one-shot and temporal relocalization approaches. Our codes are released at \href{https://github.com/zlthinker/KFNet}{https://github.com/zlthinker/KFNet}.
\end{abstract}

\vspace{-1em}
\section{Introduction}
\vspace{-0.5em}
Camera relocalization serves as the subroutine of applications including SLAM \cite{durrant2006simultaneous}, augmented reality \cite{castle2008video} and autonomous navigation \cite{royer2007monocular}. 
It estimates the 6-DoF pose of a query RGB image in a known scene coordinate system. 
Current relocalization approaches mostly focus on one-shot relocalization for a still image.
They can be mainly categorized into three classes \cite{Ding_2019_ICCV,sattler2019understanding}: (1) the relative pose regression (RPR) methods which determine the relative pose \wrt the database images \cite{balntas2018relocnet,laskar2017camera}, 
(2) the absolute pose regression (APR) methods regressing the absolute pose through PoseNet \cite{kendall2015posenet} and its variants \cite{kendall2016modelling, kendall2017geometric,walch2017image} and 
(3) the structure-based methods that establish 2D-3D correspondences with Active Search \cite{sattler2011fast, sattler2017efficient} or Scene Coordinate Regression (SCoRe) \cite{shotton2013scene} and then solve the pose by PnP algorithms \cite{gao2003complete,quan1999linear}.
Particularly, SCoRe is widely adopted recently to learn per-pixel scene coordinates from dense training data for a scene, as it can form dense and accurate 2D-3D matches even in texture-less scenes \cite{brachmann2017dsac, brachmann2018learning}.
As extensively evaluated in \cite{brachmann2017dsac, brachmann2018learning,sattler2019understanding}, the structure-based methods generally show better pose accuracy than the RPR and APR methods, because they explicitly exploit the rules of the projective geometry and the scene structures \cite{sattler2019understanding}.

Apart from one-shot relocalization, temporal relocalization with respect to video frames is also worthy of investigation.
However, almost all the temporal relocalization methods are based on PoseNet \cite{kendall2015posenet}, which, in general, even underperform the structure-based one-shot methods in accuracy. This is mainly because their accuracies are fundamentally limited by the retrieval nature of PoseNet. As analyzed in \cite{sattler2019understanding}, PoseNet based methods are essentially analogous to approximate pose estimation via image retrieval, and cannot go beyond the retrieval baseline in accuracy.

\begin{figure}[t]
\begin{center}
\includegraphics[width=1\linewidth]{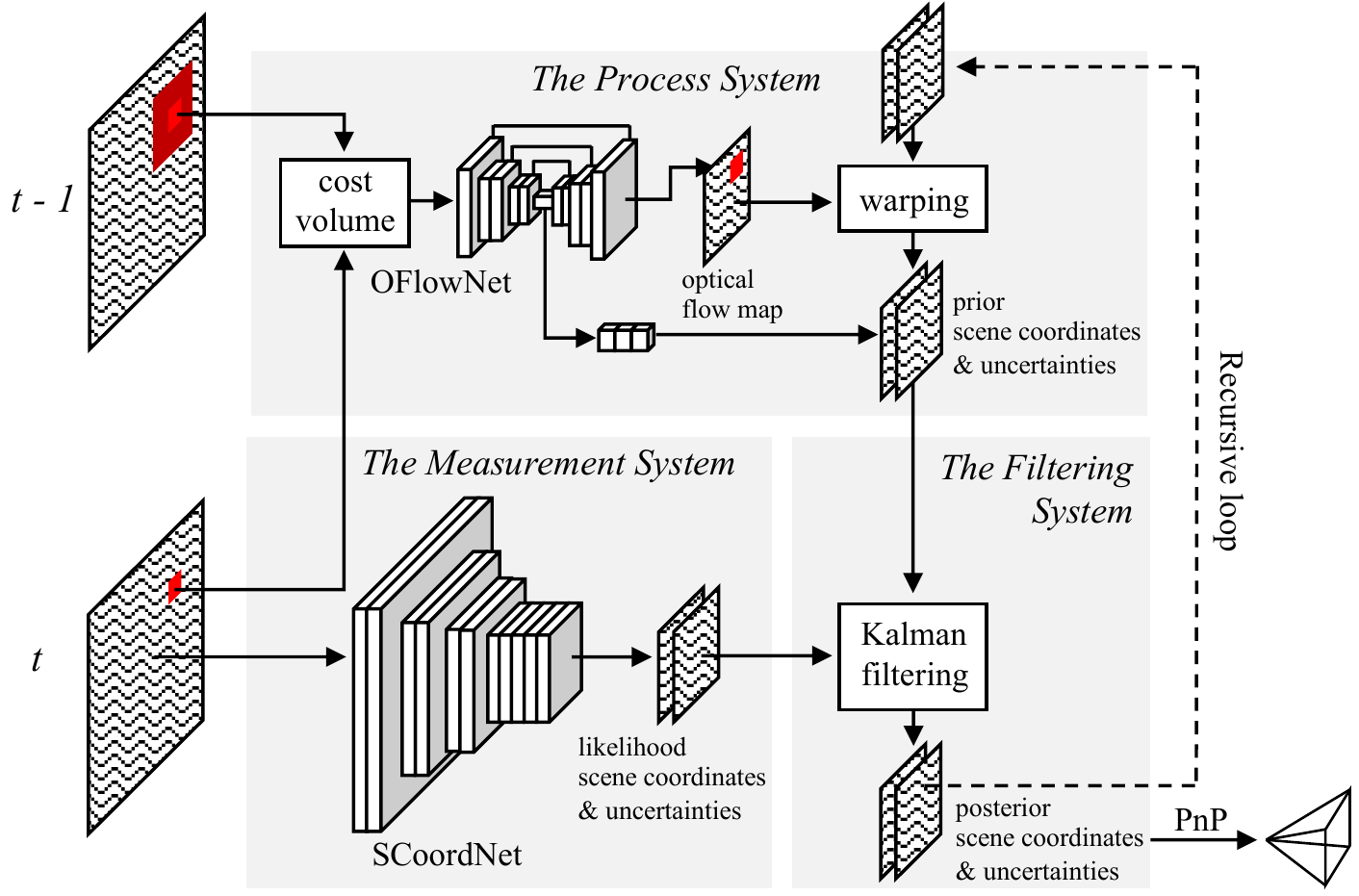}
\end{center}
   \caption{The architecture of the proposed KFNet, which is decomposed into the process, measurement and filtering systems. }
   \vspace{-1em}
\label{fig:architecture}
\end{figure}

In this work, we are motivated by the high accuracy of structure-based relocalization methods and resort to SCoRe to estimate per-pixel scene coordinates for pose computation.
Besides, we propose to extend SCoRe to the time domain in a recursive manner to enhance the temporal consistency of 2D-3D matching, thus allowing for more accurate online pose estimations for sequential images.
Specifically, a recurrent network named \textit{KFNet} is proposed in the context of Bayesian learning \cite{meinhold1983understanding} by embedding SCoRe into the Kalman filter within a deep learning framework.
It is composed of three subsystems below, as illustrated in Fig.~\ref{fig:architecture}.

\begin{itemize}[noitemsep,topsep=3pt, leftmargin=5mm]
	\item \emph{The measurement system} features a network termed \textit{SCoordNet} to derive the maximum likelihood (ML) predictions of the scene coordinates for a single image.
	\item \emph{The process system} uses \textit{OFlowNet} that models the optical flow based transition process for image pixels across time steps and yields the prior predictions of scene coordinates.
Additionally, the measurement and process systems provide uncertainty predictions \cite{novotny2017learning,kendall2016modelling} to model the noise dynamics over time.
	\item \emph{The filtering system} fuses both predictions and leads to the maximum a posteriori (MAP) estimations of the final scene coordinates.
\end{itemize}
Furthermore, we propose probabilistic losses for the three subsystems based on the Bayesian formulation of KFNet, to enable the training of either the subsystems or the full framework.
We summarize the contributions as follows.

\begin{itemize}[noitemsep,topsep=3pt, leftmargin=6mm]
	\item We are the first to extend the scene coordinate regression problem \cite{shotton2013scene} to the time domain in a learnable way for temporally-consistent 2D-3D matching.
	\item We integrate the traditional Kalman filter \cite{kalman1960new} into a recurrent CNN network (KFNet) that resolves pixel-level state inference over time-series images.
	\item KFNet bridges the existing performance gap between temporal and one-shot relocalization approaches, and achieves top accuracy on multiple relocalization benchmarks \cite{shotton2013scene, valentin2016learning, kendall2015posenet,radwan2018vlocnet++}.
	\item Lastly, for better practicality, we propose a statistical assessment tool to enable KFNet to self-inspect the potential outlier predictions on the fly.
\end{itemize}

\section{Related Works}

\smallskip\noindent\textbf{Camera relocalization.}
We categorize camera relocalization algorithms into three classes: the relative pose regression (RPR) methods, the absolute pose regression (APR) methods and the structure-based methods.

The RPR methods use a coarse-to-fine strategy which first finds similar images in the database through image retrieval \cite{torii201524,arandjelovic2016netvlad} and then computes the relative poses \wrt the retrieved images \cite{balntas2018relocnet,laskar2017camera,saha2018improved}.
They have good generalization to unseen scenes, but the retrieval process needs to match the query image against all the database images, which can be costly for time-critical applications.

The APR methods include PoseNet \cite{kendall2015posenet} and its variants \cite{kendall2016modelling, kendall2017geometric, walch2017image} which learn to regress the absolute camera poses from the input images through a CNN.
They are simple and efficient, but generally fall behind the structure-based methods in terms of accuracy, as validated by \cite{brachmann2017dsac, brachmann2018learning,sattler2019understanding}.
Theoretically, \cite{sattler2019understanding} explains that PoseNet-based methods are more closely related to image retrieval than to accurate pose estimation via 3D geometry.

The structure-based methods explicitly establish the correspondences between 2D image pixels and 3D scene points and then solve camera poses by PnP algorithms \cite{gao2003complete, quan1999linear, lepetit2009epnp}.
Traditionally, correspondences are searched by matching the patch features against Structure from Motion (SfM) tracks via Active Search \cite{sattler2011fast, sattler2017efficient} and its variants \cite{li2010location, choudhary2012visibility, Liu_2017_ICCV,Sarlin_2019_CVPR}, which can be inefficient and fragile in texture-less scenarios.
Recently, the correspondence problem is resolved by predicting the scene coordinates for pixels by training random forests \cite{shotton2013scene,valentin2015exploiting,massiceti2017random} or CNNs \cite{brachmann2017dsac, brachmann2018learning, li2018scene,esac} with ground truth scene coordinates, which is referred to as Scene Coordinate Regression (SCoRe).

Besides one-shot relocalization, some works have extended PoseNet to the time domain to address temporal relocalization.
VidLoc \cite{clark2017vidloc} performs offline and batch relocalization for fixed-length video-clips by BLSTM \cite{schuster1997bidirectional}.
Coskun \etal refine the pose dynamics by embedding LSTM units in the Kalman filters \cite{coskun2017long}.
VLocNet \cite{valada2018deep} and VLocNet++ \cite{radwan2018vlocnet++} propose to learn pose regression and the visual odometry jointly.
LSG \cite{xue2019local} combines LSTM with visual odometry to further exploit the spatial-temporal consistency.
Since all the methods are extensions of PoseNet, their accuracies are fundamentally limited by the retrieval nature of PoseNet, following the analysis of \cite{sattler2019understanding}.

\smallskip\noindent\textbf{Temporal processing.}
When processing time-series image data, ConvLSTM \cite{xingjian2015convolutional} is a standard way of modeling the spatial correlations of local contexts through time \cite{valipour2017recurrent,Luo_2018_CVPR,Lai_2018_ECCV}. 
However, some works have pointed out that the implicit convolutional modeling is less suited to discovering the pixel associations between neighboring frames, especially when pixel-level accuracy is desired \cite{Ilg_2017_CVPR,Nilsson_2018_CVPR}.
Therefore, in later works, the optical flow is highlighted as a more explicit way of delineating the pixel correspondences across sequential steps \cite{pfister2015flowing}. 
For example, \cite{pfister2015flowing, Kim_2018_ECCV, Lai_2018_ECCV, Song_2017_CVPR,Nilsson_2018_CVPR} commonly predict the optical flow fields to guide the feature map warping across time steps. 
Then, the warped features are fused by weighting \cite{Zhu_2017_ICCV,Zhu_2018_CVPR} or pooling \cite{nguyen2018weakly,pfister2015flowing} to aggregate the temporal knowledge.
In this work, we follow the practice of flow-guided warping, but the distinction from previous works is that we propose to fuse the predictions by leveraging Kalman filter principles \cite{meinhold1983understanding}.

\section{Bayesian Formulation} \label{sec:formulation}
This section presents the Bayesian formulation of recursive scene coordinate regression in the time domain for temporal camera relocalization.
Based on the formulation, the proposed KFNet is built and the probabilistic losses are defined in Sec. \ref{sec:SCoRe} $\sim$ \ref{sec:KF}. Notations used below have been summarized in Table~\ref{table:notation} for quick reference.

Given a stream of RGB images up to time $t$, \ie, $\mathcal{I}_t = \{\mathbf{I}_1, ..., \mathbf{I}_{t-1}, \mathbf{I}_t\} $, 
our aim is to predict the latent state for each frame, \ie, the scene coordinate map, which is then used for pose computation. We denote the map as $\boldsymbol{\theta}_t \in \mathbb{R}^{N \times 3}$, where $N$ is the pixel number. By imposing the Gaussian noise assumption on the states, the state $\boldsymbol{\theta}_t$ conditioned on $\mathcal{I}_t$ follows an unknown Gaussian distribution:
\begin{equation}
\boldsymbol{\theta}_t^{+}  \overset{\underset{\mathrm{def}}{}}{=} (\boldsymbol{\theta}_t | \mathcal{I}_t ) \sim \mathcal{N}(\hat{\boldsymbol{\theta}}_t, \mathbf{\Sigma}_t),
\end{equation}
where $\hat{\boldsymbol{\theta}}_t$ and $\mathbf{\Sigma}_t$ are the expectation and covariance to be determined.
Under the routine of Bayesian theorem, the posterior probability of $\boldsymbol{\theta}_t$ can be factorized as
\begin{equation} \label{eq:posterior}
P(\boldsymbol{\theta}_t | \mathcal{I}_t) \propto P(\boldsymbol{\theta}_t  | \mathcal{I}_{t-1}) P(\mathbf{I}_t | \boldsymbol{\theta}_t ,\mathcal{I}_{t-1}),
\end{equation}
where $\mathcal{I}_t = \mathcal{I}_{t-1} \cup \{\mathbf{I}_t \}$.

\iffalse
\footnote{$\mathbb{S}_{++}^{N}$ denotes the set of N-d positive definite matrices.}
\fi

The first factor $P(\boldsymbol{\theta}_t  | \mathcal{I}_{t-1})$ of the right hand side (RHS) of Eq.~\ref{eq:posterior} indicates the prior belief about $\boldsymbol{\theta}_t $ obtained from time $t-1$ through a \textit{process system}. Provided that no occlusions or dynamic objects occur, the consecutive coordinate maps can be approximately associated by a linear \textbf{process equation} describing their pixel correspondences, wherein
\begin{equation} \label{eq:transition}
\boldsymbol{\theta}_t = \mathbf{G}_t \boldsymbol{\theta}_{t-1} + \mathbf{w}_{t},
\end{equation}
with $\mathbf{G}_t \in \mathbb{R}^{N \times N}$ being the sparse state transition matrix given by the optical flow fields from time $t-1$ to $t$, and $\mathbf{w}_{t} \sim \mathcal{N}(\mathbf{0}, \mathbf{W}_t)$, $\mathbf{W}_t \in \mathbb{S}_{++}^{N}$\footnote{$\mathbb{S}_{++}^{N}$ denotes the set of N-dimensional positive definite matrices.} being the \textit{process noise}.
Given $\mathcal{I}_{t-1}$, we already have the probability statement that $(\boldsymbol{\theta}_{t-1} | \mathcal{I}_{t-1} ) \sim \mathcal{N}(\hat{\boldsymbol{\theta}}_{t-1}, \mathbf{\Sigma}_{t-1})$.
Then the prior estimation of $\boldsymbol{\theta}_t$ from time $t-1$ can be expressed as
\begin{equation} \label{eq:distribution_transition}
\boldsymbol{\theta}_t^{-}  \overset{\underset{\mathrm{def}}{}}{=} (\boldsymbol{\theta}_t | \mathcal{I}_{t-1} ) \sim   \mathcal{N}(\hat{\boldsymbol{\theta}}_t^{-}, \mathbf{R}_t) ,
\end{equation}
where $\hat{\boldsymbol{\theta}}_t^{-} = \mathbf{G}_t \hat{\boldsymbol{\theta}}_{t-1}$, $\mathbf{R}_t = \mathbf{G}_t \mathbf{\Sigma}_{t-1} \mathbf{G}_t^T + \mathbf{W}_t$.

\begin{table}[]
\resizebox{\linewidth}{!}
{
\bgroup
\begin{tabular}{c|c|clc}
\Xhline{2\arrayrulewidth}
Module    & inputs & \multicolumn{2}{c}{outputs} \\ \hline \hline
\begin{tabular}[c]{@{}c@{}}The \\process \\system   \end{tabular} &    
\begin{tabular}[c]{@{}c@{}}
$\hat{\boldsymbol{\theta}}_{t-1}$\\
$\mathbf{\Sigma}_{t-1}$ \\
$\mathbf{I}_{t-1}$ \\
$\mathbf{I}_t$    \end{tabular} &
\begin{tabular}[c]{@{}c@{}}
$\mathbf{G}_{t}$ \\ 
$\mathbf{W}_{t}$ \\
$\hat{\boldsymbol{\theta}}_t^{-} = \mathbf{G}_t \hat{\boldsymbol{\theta}}_{t-1}$ \\
$\mathbf{R}_t = \mathbf{G}_t \mathbf{\Sigma}_{t-1} \mathbf{G}_t^T + \mathbf{W}_t$  
\end{tabular}  &

\begin{tabular}{l}
\textit{- transition matrix}\\ 
\textit{- process noise covariance}\\
\textit{- prior state mean}\\
\textit{- prior state covariance}
\end{tabular}  
  
 \\ \hline
\begin{tabular}[c]{@{}c@{}}The \\measurement \\system \end{tabular}&   $\mathbf{I}_t$     &   
\begin{tabular}[c]{@{}c@{}}
$\mathbf{z}_t$  \\ 
$\mathbf{V}_t$  \\
\\
\end{tabular} &

\begin{tabular}{l}
\textit{- state observations}\\ 
\textit{- measurement noise } \\
\textit{\;\,\,covariance} \\ \end{tabular}
 \\ \hline

\begin{tabular}[c]{@{}c@{}} The \\filtering \\system  \end{tabular}  & 
\begin{tabular}[c]{@{}c@{}}
$\hat{\boldsymbol{\theta}}_t^{-}$\\ 
$\mathbf{z}_t$ \\
$\mathbf{R}_t$ \\ 
$\mathbf{V}_t$      
\end{tabular} &
\begin{tabular}[c]{@{}c@{}}  
$\mathbf{e}_t = \mathbf{z}_t -  \hat{\boldsymbol{\theta}}_t^{-}$\\ 
$\mathbf{K}_t =\frac{\mathbf{R}_t}{\mathbf{V}_t + \mathbf{R}_t} $\\
$\hat{\boldsymbol{\theta}}_t = \hat{\boldsymbol{\theta}}_t^{-} + \mathbf{K}_t \mathbf{e}_t$ \\
$\mathbf{\Sigma}_t  =  \mathbf{R}_t (\mathbf{I} - \mathbf{K}_t)$ 
\end{tabular}  &
\begin{tabular}{l}  
\textit{- innovation}\\ 
\textit{- Kalman gain}\\
\textit{- posterior state mean} \\
\textit{- posterior state covariance} \\
\end{tabular}

\\ \Xhline{2\arrayrulewidth}
\end{tabular}
\egroup
}
\caption{The summary of variables and notations used in the Bayesian formulation of KFNet.}
\vspace{-1em}
\label{table:notation}
\end{table}

The second factor $P(\mathbf{I}_t | \boldsymbol{\theta}_t ,\mathcal{I}_{t-1})$ of the RHS of Eq.~\ref{eq:posterior} describes the likelihood of image observations at time $t$ made through a \textit{measurement system}. 
The system models how $\mathbf{I}_t$ is derived from the latent states $\boldsymbol{\theta}_t$, formally $\mathbf{I}_t = \mathbf{h}(\boldsymbol{\theta}_t)$. 
However, the high nonlinearity of $\mathbf{h}(\cdot)$ makes the following computation intractable.
Alternatively, we map $\mathbf{I}_t$ to $\mathbf{z}_t \in \mathbb{R}^{N \times 3}$ via a nonlinear function inspired by \cite{coskun2017long}, so that the system can be approximately expressed by a linear \textbf{measurement equation}:
\begin{equation} \label{eq:observation}
\textbf{z}_t  = \boldsymbol{\theta}_t + \mathbf{v}_t,
\end{equation}
where $\mathbf{v}_t \sim \mathcal{N}(\mathbf{0}, \mathbf{V}_t)$, $\mathbf{V}_t \in \mathbb{S}_{++}^{N}$ denotes the \textit{measurement noise}, and $\textbf{z}_t$ can be interpreted as the noisy observed scene coordinates.
In this way, the likelihood can be re-written as $P(\mathbf{z}_t | \boldsymbol{\theta}_t ,\mathcal{I}_{t-1})$ by substituting $\mathbf{z}_t$ for $\mathbf{I}_t$. 

Let $\mathbf{e}_t $ denote the residual of predicting $\mathbf{z}_t$ from time $t-1$; thus
\begin{equation} \label{eq:predict_error}
\mathbf{e}_t = \mathbf{z}_t -  \hat{\boldsymbol{\theta}}_t^{-} = \mathbf{z}_t - \mathbf{G}_t \hat{\boldsymbol{\theta}}_{t-1}.
\end{equation}
Since $\mathbf{G}_t$ and $\hat{\boldsymbol{\theta}}_{t-1}$ are all known, observing $\mathbf{z}_t$ is equivalent to observing $\mathbf{e}_t$. Hence, the likelihood $P(\mathbf{z}_t | \boldsymbol{\theta}_t ,\mathcal{I}_{t-1})$ can be rewritten as $P(\mathbf{e}_t | \boldsymbol{\theta}_t ,\mathcal{I}_{t-1})$. Substituting Eq.~\ref{eq:observation} into Eq.~\ref{eq:predict_error}, we have $\mathbf{e}_t = \boldsymbol{\theta}_t -  \hat{\boldsymbol{\theta}}_t^{-} + \mathbf{v}_t$, so that the likelihood can be described by 
\begin{equation}  \label{eq:distribution_measurement}
(\mathbf{e}_t | \boldsymbol{\theta}_t ,\mathcal{I}_{t-1}) \sim \mathcal{N}(\boldsymbol{\theta}_t -  \hat{\boldsymbol{\theta}}_t^{-}, \mathbf{V}_t).
\end{equation}

Based on the theorems in multivariate statistics \cite{anderson1958introduction, meinhold1983understanding}, combining the two distributions \ref{eq:distribution_transition} \& \ref{eq:distribution_measurement} gives the bivariate normal distribution:
\begin{equation}   \label{eq:bivariate}
\left[  \left. \begin{pmatrix} \boldsymbol{\theta}_t \\ \mathbf{e}_t \end{pmatrix} \right\vert \mathcal{I}_{t-1} \right] \sim \mathcal{N} \left[ \begin{pmatrix} \hat{\boldsymbol{\theta}}_t^{-} \\ \mathbf{0} \end{pmatrix} , \begin{pmatrix}  \mathbf{R}_t & \mathbf{R}_t \\ \mathbf{R}_t & \mathbf{R}_t + \mathbf{V}_t \end{pmatrix} \right].
\end{equation}
Making $\mathbf{e}_t$ the conditioning variable, \textit{the filtering system} gives the posterior distribution that writes
\begin{align} \label{eq:posterior_derived}
\begin{split}
\boldsymbol{\theta}_t^+ 
& \overset{\underset{\mathrm{def}}{}}{=} (\boldsymbol{\theta}_t | \mathcal{I}_t)
= (\boldsymbol{\theta}_t | \mathbf{e}_t, \mathcal{I}_{t-1})  \sim \mathcal{N}(\hat{\boldsymbol{\theta}}_t, \mathbf{\Sigma}_t)
\\ & \sim \mathcal{N}(\hat{\boldsymbol{\theta}}_t^{-} + \mathbf{K}_t \mathbf{e}_t    ,    \mathbf{R}_t (\mathbf{I} - \mathbf{K}_t) ),
\end{split}
\end{align}
where $\mathbf{K}_t = \frac{\mathbf{R}_t}{\mathbf{V}_t + \mathbf{R}_t}$ is conceptually referred to as the \textit{Kalman gain} and $ \mathbf{e}_t $ as the \textit{innovation}\footnote{The derivation of Eqs.~\ref{eq:bivariate} \& \ref{eq:posterior_derived} is shown in Appendix~\ref{sec:bayesian_derivation}.} \cite{meinhold1983understanding,grewal2011kalman}.
\iffalse
The posterior solution of $\boldsymbol{\theta}_t$ is provably optimal due to the linear least-square nature of the Kalman filter \cite{grewal2011kalman}.
\fi

As shown in Fig.~\ref{fig:architecture}, the inference of the posterior scene coordinates $\hat{\boldsymbol{\theta}}_t$ and covariance $\mathbf{\Sigma}_t$ for image pixels proceeds recursively as the time $t$ evolves, 
which are then used for online pose determination.
Specifically, the pixels with variances greater than $\lambda$ are first excluded as outliers.
Then, a RANSAC+P3P \cite{gao2003complete} solver is applied to compute the initial camera pose from the 2D-3D correspondences, followed by a nonlinear optimization for pose refinement.

\section{The Measurement System} \label{sec:SCoRe}
The measurement system is basically a generative model explaining how the observations $\mathbf{z}_t$ are generated from the latent scene coordinates $\boldsymbol{\theta}_t$, as expressed in Eq.~\ref{eq:observation}. 
Then, the remaining problem is to learn the underlying mapping from $\mathbf{I}_t$ to $\mathbf{z}_t$.
This is similar to the SCoRe task \cite{shotton2013scene, brachmann2017dsac, brachmann2018learning}, but differs in the constraint about $\mathbf{z}_t$ imposed by Eq.~\ref{eq:observation}.
Below, the architecture of SCoordNet is first introduced, which outputs the scene coordinate predictions, along with the uncertainties, to model the measurement noise $\mathbf{v}_t$.
Then, we define the probabilistic loss based on the likelihood $P(\mathbf{z}_t | \boldsymbol{\theta}_t, \mathcal{I}_{t-1} )$ of the measurement system.

\begin{figure}[t]
\begin{center}
\includegraphics[width=1\linewidth]{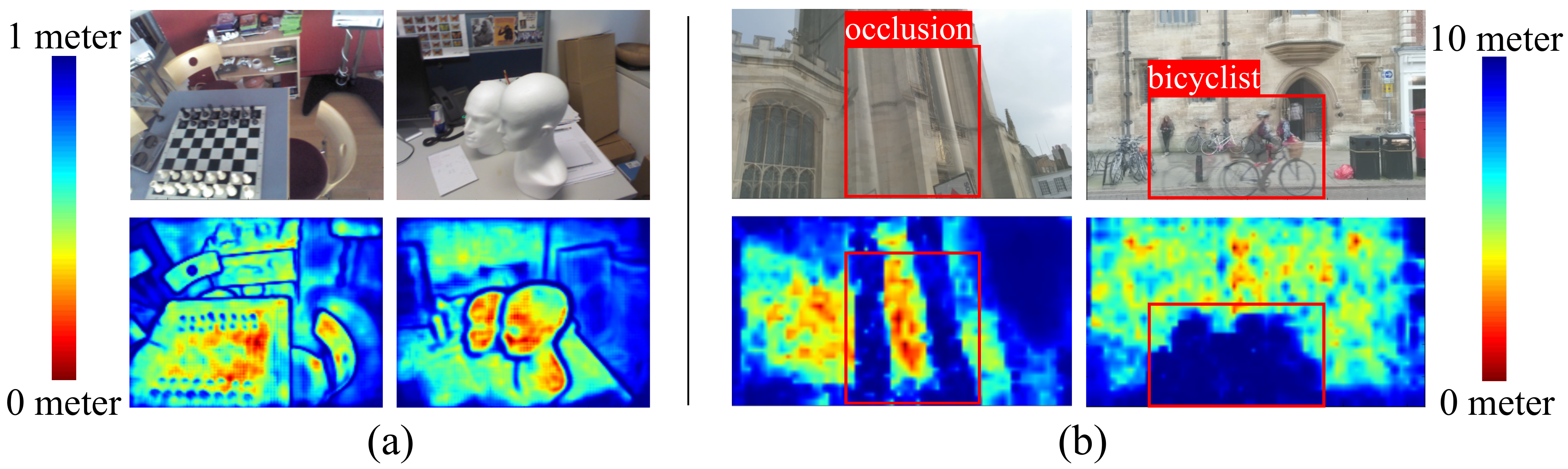}
\end{center}
   \caption{The visualization of uncertainties which model the measurement noise and the process noise.
(a) SCoordNet predicts larger uncertainties from single images over the object boundaries where larger errors occur. 
(b) OFlowNet gives larger uncertainties from the consecutive images (overlaid) over the areas where occlusions or dynamic objects appear.}
\vspace{-1em}
\label{fig:uncertainty}
\end{figure}

\subsection{Architecture}

SCoordNet shares the similar fully convolutional structure to \cite{brachmann2018learning}, as shown in Fig.~\ref{fig:architecture}. However, it is far more lightweight, with parameters fewer than one eighth of \cite{brachmann2018learning}.
It encompasses twelve $3\times3$ convolution layers, three of which use a stride of $2$ to downsize the input by a factor of $8$. 
ReLU follows each layer except the last one.
To simplify computation and avoid the risk of over-parameterization,
we postulate the isotropic covariance of the multivariate Gaussian measurement noise, \ie, $\mathbf{V}_{(i)} = {{v}_{(i)}}^2 \mathbb{I}_3$ for each pixel $\mathbf{p}_i$, where $\mathbb{I}_3$ denotes the $3\times3$ identity matrix.
The output thus has a channel of $4$, comprising $3$-d scene coordinates and a $1$-d uncertainty measurement.

\subsection{Loss}
%\paragraph{Probabilistic loss}
According to Eq.~\ref{eq:observation}, the latent scene coordinates $\boldsymbol{\theta}_{(i)}$ of pixel $\mathbf{p}_i$ should follow the distribution $\mathcal{N}(\mathbf{z}_{(i)}, {v}_{(i)}^2 \mathbb{I}_3)$.
Taking the negative logarithm of the probability density function (PDF) of $\boldsymbol{\theta}_{(i)}$, we define the loss based on the likelihood which gives rise to the maximum likelihood (ML) estimation for each pixel in the form \cite{kendall2016modelling}:
\begin{equation} \label{eq:likelihood_loss}
\mathcal{L}_{likelihood} = \sum_{i=1}^N \left(3 \log {v}_{(i)} + \frac{\| \mathbf{z}_{(i)} - \mathbf{y}_{(i)}  \|_2^2}{ 2  {{v}_{(i)}}^2 }  \right),
\end{equation}
with $\mathbf{y}_{(i)}$ being the groundtruth label for $\boldsymbol{\theta}_{(i)}$.
For numerical stability, we use logarithmic variance for the uncertainty measurements in practice, \ie, ${s}_{(i)} = \log {{v}_{(i)}}^2$.

Including uncertainty learning in the loss formulation allows one to quantify the prediction errors stemming not just from the intrinsic noise in the data but also from the defined model \cite{do2007gaussian}.
For example, at the boundary with depth discontinuity, a sub-pixel offset would cause an abrupt coordinate shift which is hard to model.
SCoordNet would easily suffer from a significant magnitude of loss in such cases.
It is sensible to automatically downplay such errors during training by weighting with the uncertainty measurements.
Fig.~\ref{fig:uncertainty}(a) illustrates the uncertainty predictions in such cases.

\iffalse
\paragraph{Reprojection loss}

Besides, we supply the loss definition with the reprojection loss of predicted coordinates. We use the definition of angle-based loss \cite{li2018scene}, which writes
\begin{equation}
\mathcal{L}_r = \sum_i  \left\|  \frac{\mathbf{T}(\mathbf{z}_{(i)}) }{ \| \mathbf{T}(\mathbf{z}_{(i)} ) \|_2} - \frac{\mathbf{K}^{-1} \tilde{\mathbf{p}}_{(i)} } { \| \mathbf{K}^{-1} \tilde{\mathbf{p}}_{(i)} \|_2}   \right\|_2^2,
\end{equation}
where $\mathbf{T}(\cdot)$ transforms the global coordinates to the local camera frame by camera extrinsics, $\mathbf{K}$ is the intrinsic camera matrix and $\tilde{\mathbf{p}}_{(i)}$ is the homogeneous coordinates of pixel $\mathbf{p}_i$. Using the angle error between two rays avoids the degeneration case when points locate behind the camera center illegitimately, as opposed to the pixel error used in \cite{brachmann2018learning}.

The final loss for training SCoordNet adds up the two loss terms as
\begin{equation}
\mathcal{L}_{coord} = \mathcal{L}_p + \alpha \mathcal{L}_r.
\end{equation}
\fi

\section{The Process System} \label{sec:flow}
The process system models the transition process of pixel states from time $t-1$ to $t$, as described by the process equation of Eq.~\ref{eq:transition}.
Herein, first, we propose a cost volume based network, OFlowNet, to predict the optical flows and the process noise covariance jointly for each pixel.
Once the optical flows are determined, Eq.~\ref{eq:transition} is equivalent to the flow-guided warping from time $t-1$ towards $t$, as commonly used in \cite{pfister2015flowing, Kim_2018_ECCV, Lai_2018_ECCV, Song_2017_CVPR,Nilsson_2018_CVPR}.
Second, after the warping, the prior distribution of the states, \ie, $\boldsymbol{\theta}_t^{-} \sim   \mathcal{N}(\hat{\boldsymbol{\theta}}_t^{-}, \mathbf{R}_t)$ of Eq.~\ref{eq:distribution_transition}, can be evaluated.
We then define the probabilistic loss based on the prior to train OFlowNet.

\begin{figure}[t]
\begin{center}
\includegraphics[width=0.95\linewidth]{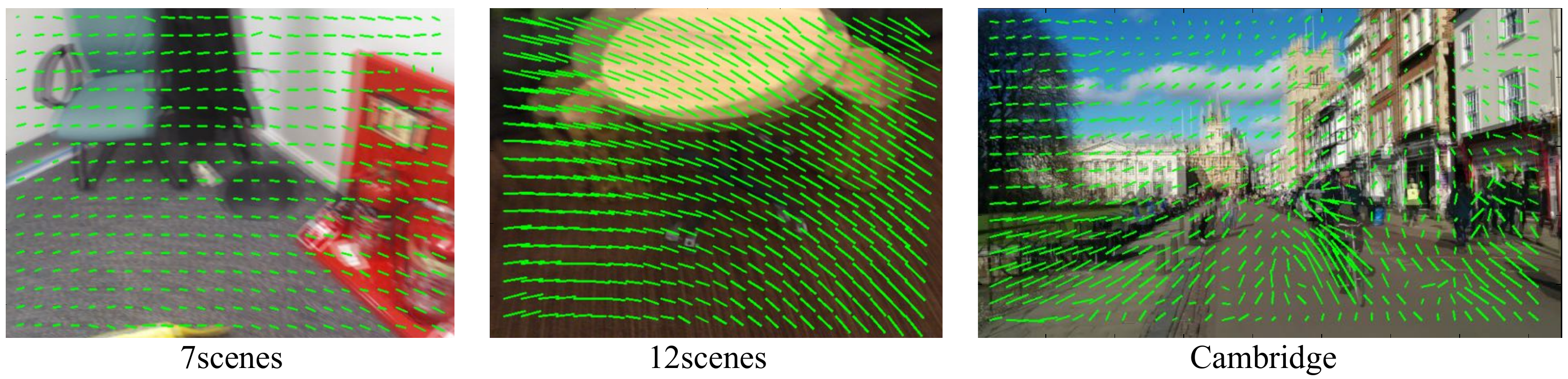}
\end{center}
  \caption{Sample optical flows predicted by OFlowNet over consecutive images (overlaid) of three different datasets \cite{shotton2013scene,valentin2016learning,kendall2015posenet}.}
  \vspace{-1em}
\label{fig:optical_flow}
\end{figure}

\subsection{Architecture}

OFlowNet is composed of two components: the cost volume constructor and the flow estimator.

The cost volume constructor first extracts features from the two input images $\mathbf{I}_{t-1}$ and $\mathbf{I}_t$ respectively through seven $3\times3$ convolutions, three of which have a stride of $2$ . 
The output feature maps $\mathbf{F}_{t-1}$ and $\mathbf{F}_t$ have a spatial size of one-eighth of the inputs and a channel number of $c$.
Then, we build up a cost volume $\mathbf{C}_i \in \mathbb{R}_{+}^{w \times w \times c}$ for each pixel $\mathbf{p}_i$ of the feature map $\mathbf{F}_t$, so that
\begin{equation}
\mathbf{C}_i(\mathbf{o}) = \left| \frac{\mathbf{F}_t(\mathbf{p}_i)}{  \| \mathbf{F}_t(\mathbf{p}_i) \|_2} -  \frac{\mathbf{F}_{t-1}(\mathbf{p}_i + \mathbf{o}) }{ \| \mathbf{F}_{t-1}(\mathbf{p}_i + \mathbf{o})  \|_2}  \right|,
\end{equation} 
where $w$ is the size of the search window which corresponds to $8w$ pixels in the full-resolution image, and $\mathbf{o} \in \{-w/2, ..., w/2 \}^2$ is the spatial offset.
We apply L2-normalization to the feature maps along the channel dimension before differentiation, as in \cite{Xu_2017_CVPR, Lu_2017_ICCV}.

The following flow estimator operates over the cost volumes for flow inference. 
We use a U-Net with skip connections \cite{ronneberger2015u} as shown in Fig.~\ref{fig:architecture}, which first subsamples the cost volume by a factor of $8$ for an enlarged receptive field and then upsamples it to the original resolution. The output is a $w \times w \times 1$ unbounded confidence map for each pixel.
Related works usually attain flows by hard assignment based on the matching cost encapsulated by the cost volumes \cite{Xu_2017_CVPR,sun2018pwc}. 
However, it would cause non-differentiability in later steps where the optical flows are to be further used for spatial warping.
Thus, we pass the confidence map through the differentiable \textit{spatial softmax} operator \cite{finn2015learning} to compute the optical flow as the expectation of the pixel offsets inside the search window. Formally,
\begin{equation}
\hat{\mathbf{o}} \overset{\underset{\mathrm{def}}{}}{=} \mathrm{E}(\mathbf{o}) = \sum_{\mathbf{o}} \textrm{softmax}(f_\mathbf{o}) \cdot \mathbf{o},
\end{equation}
where $f_\mathbf{o}$ is the confidence at offset $\mathbf{o}$.
To fulfill the process noise modeling, \ie, $\mathbf{w}_t$ in Eq.~\ref{eq:transition}, we append three fully connected layers after the bottleneck of the U-Net to regress the logarithmic variance, as shown in Fig.~\ref{fig:architecture}.
Sample optical flow predictions are visualized in Fig.~\ref{fig:optical_flow}.

\subsection{Loss}

Once the optical flows are computed, the state transition matrix $\mathbf{G}_t$ of Eq.~\ref{eq:transition} can be evaluated.
We then complete the linear transition process of Eq.~\ref{eq:transition} by warping the scene coordinate map and uncertainty map from time $t-1$ towards $t$ through bilinear warping \cite{Zhou_2017_CVPR}.
Let $\hat{\boldsymbol{\theta}}_{(i)}^-$ and ${\sigma_{(i)}^-}^2$ be the warped scene coordinates and Gaussian variance, and ${{w}_{(i)}}^2$ be the Gaussian variance of the process noise of pixel $\mathbf{p}_i$ at time $t$.
Then, the prior coordinates of $\mathbf{p}_i$, denoted as $\boldsymbol{\theta}_{(i)}^-$, should follow the distribution
\begin{equation}
\boldsymbol{\theta}_{(i)}^- \sim \mathcal{N}(\hat{\boldsymbol{\theta}}_{(i)}^-, {{r}_{(i)} }^2 \mathbb{I}_3),
\end{equation}
where ${{r}_{(i)}}^2 = {\sigma_{(i)}^-}^2 + {{w}_{(i)}}^2$.
Taking the negative logarithm of the PDF of $\boldsymbol{\theta}_{(i)}^-$, 
we get the loss of the process system as 
\begin{equation}  \label{eq:prior_loss}
\mathcal{L}_{prior} = \sum_{i=1}^N \left(3 \log {{r}_{(i)}} + \frac{\| \hat{\boldsymbol{\theta}}_{(i)}^- - \mathbf{y}_{(i)}  \|_2^2}{ 2  {{r}_{(i)}}^2 }  \right).
\end{equation}
It is noteworthy that the loss definition uses the prior distribution of $\boldsymbol{\theta}_{(i)}^-$ to provide the weak supervision for training OFlowNet, with no recourse to the optical flow labeling.

One issue with the proposed process system is that it assumes no occurrence of occlusions or dynamic objects which are two outstanding challenges for tracking problems \cite{koller1994robust, zou2013coslam}.
Our process system partially addresses the issue by giving the uncertainty measurements of the process noise.
As shown in Fig.~\ref{fig:uncertainty}(b), OFlowNet generally produces much larger uncertainty estimations for the pixels from occluded areas and dynamic objects.
This helps to give lower weights to these pixels that have incorrect flow predictions in the loss computation.

\section{The Filtering System} \label{sec:KF}
The measurement and process systems in the previous two sections have derived the likelihood and prior estimations of the scene coordinates $\boldsymbol{\theta}_t$, respectively.
The filtering system aims to fuse both of them based on Eq.~\ref{eq:posterior_derived} to yield the posterior estimation.

\subsection{Loss}
For a pixel $\mathbf{p}_i$ at time $t$, $\mathcal{N}(\mathbf{z}_{(i)}, {{v}_{(i)}}^2  \mathbb{I}_3)$ and $\mathcal{N}(\hat{\boldsymbol{\theta}}_{(i)}^-, {{r}_{(i)} }^2  \mathbb{I}_3)$
are respectively the likelihood and prior distributions of its scene coordinates.
Putting the variables in Eqs.~\ref{eq:predict_error} \& \ref{eq:posterior_derived}, we evaluate the innovation and the Kalman gain at pixel $\mathbf{p}_i$ as
\begin{equation}
\mathbf{e}_{(i)} = \mathbf{z}_{(i)} - \hat{\boldsymbol{\theta}}_{(i)}^-, \;\;
 \text{and} \;\; {k}_{(i)} = \frac{{{r}_{(i)} }^2}{{{v}_{(i)}}^2 + {{r}_{(i)} }^2}.
\end{equation}
Imposing the linear Gaussian postulate of the Kalman filter, the fused scene coordinates of $\mathbf{p}_i$ with the least square error follow the posterior distribution below \cite{meinhold1983understanding} :
\begin{equation}
\boldsymbol{\theta}_{(i)}^+   \sim \mathcal{N}(\hat{\boldsymbol{\theta}}_{(i)}^+, {\sigma_{(i)}}^2 \mathbb{I}_3),
\end{equation}
where $\hat{\boldsymbol{\theta}}_{(i)}^+ = \hat{\boldsymbol{\theta}}_{(i)}^- + {k}_{(i)} \mathbf{e}_{(i)} $ and $ {\sigma_{(i)}}^2 = {{r}_{(i)} }^2 (1 - {k}_{(i)})$.
Hence, the Kalman filtering system is parameter-free, with the loss defined based on the posterior distribution:
\begin{equation} 
\mathcal{L}_{posterior} = \sum_{i=1}^N \left(3 \log {\sigma_{(i)}} + \frac{\| \hat{\boldsymbol{\theta}}_{(i)}^+ - \mathbf{y}_{(i)}  \|_2^2}{ 2  {\sigma_{(i)}}^2 }  \right),
\end{equation}
which is then added to the full loss that allows the end-to-end training of KFNet as below:
\begin{equation}  \label{eq:full_loss}
\mathcal{L}_{full} = \tau_1 \mathcal{L}_{likelihood} + \tau_2 \mathcal{L}_{prior}  + \tau_3 \mathcal{L}_{posterior}.
\end{equation}

\begin{figure}[t]
\begin{center}
\includegraphics[width=0.65\linewidth]{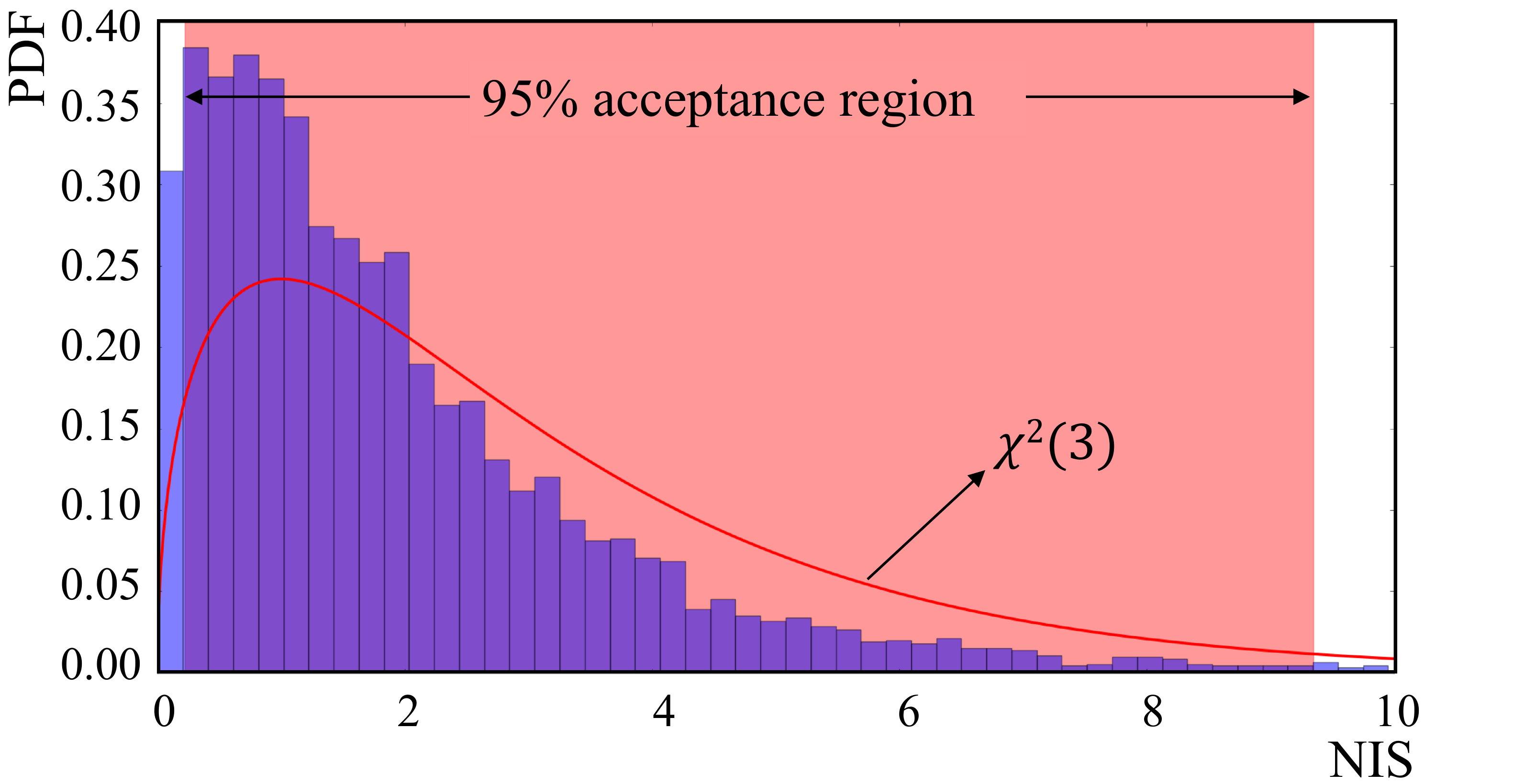}
\end{center}
   \caption{The illustration of NIS testing for the filtering system. The histogram draws the exemplar distribution of the Normalized Innovation Squared (NIS) values of the Kalman filter. The red curve denotes the PDF of the 3-DoF Chi-squared distribution $\chi^2(3)$. NIS testing works by filtering out the inconsistent predictions whose NIS values locate out of the $95\%$ acceptance region (red shaded) of $\chi^2(3)$.}
\vspace{-1em}
\label{fig:chi-square}
\end{figure}

\subsection{Consistency Examination} \label{sec:consistency}

In practice, the filter could behave incorrectly due to the outlier estimations caused by the erratic scene coordinate regression or a failure of flow tracking. 
This would induce accumulated state errors in the long run.
Therefore, we use the statistical assessment tool, \textit{Normalized Innovation Squared (NIS)} \cite{bar2004estimation}, to filter the inconsistent predictions during inference.

\begin{table*}[t]
\begin{threeparttable}
{
%\centering
\resizebox{\linewidth}{!}
{
\begin{tabular}{c|c?c|c|c|c|c?c|c|c|c|c}
\Xhline{2\arrayrulewidth}
     &   & \multicolumn{5}{c?}{One-shot Relocalization}                                                                                                  & \multicolumn{5}{c}{Temporal Relocalization}    \\ \hline
                   & scenes           
                   & \begin{tabular}[c]{@{}c@{}}MapNet \\ \cite{Brahmbhatt_2018_CVPR}\end{tabular} 
                   & \begin{tabular}[c]{@{}c@{}}CamNet \\ \cite{Ding_2019_ICCV}\end{tabular} 
                   & \begin{tabular}[c]{@{}c@{}}Active \\ Search \cite{sattler2017efficient} \end{tabular}      
                   & \begin{tabular}[c]{@{}c@{}}DSAC++ \\ \cite{brachmann2018learning}  \end{tabular} 
                   & \begin{tabular}[c]{@{}c@{}}SCoordNet \\ (Ours)\end{tabular}    
                   & \begin{tabular}[c]{@{}c@{}}VidLoc \\ \cite{clark2017vidloc}  \end{tabular} 
                   & \begin{tabular}[c]{@{}c@{}}LSTM-KF \\ \cite{coskun2017long}  \end{tabular}  
                   & \begin{tabular}[c]{@{}c@{}}VLocNet++ \\ \cite{radwan2018vlocnet++}  \end{tabular}  
                   & \begin{tabular}[c]{@{}c@{}}LSG \\ \cite{xue2019local}  \end{tabular}  
                   & \begin{tabular}[c]{@{}c@{}}KFNet \\ (Ours)\end{tabular} 
                  \\ \hline\hline
\multirow{8}{*}{\rotatebox[origin=c]{90}{7scenes}}   
& chess            & 0.08m, 3.25\textdegree    & 0.04m, 1.73\textdegree                                               & 0.04m, 1.96\textdegree                                                 & 0.02m, 0.5\textdegree   & 0.019m, 0.63\textdegree & 0.18m, - & 0.33m, 6.9\textdegree    & 0.023m,1.44\textdegree  &  0.09m, 3.28\textdegree & 0.018m, 0.65\textdegree \\ \cline{2-12} 
& fire             & 0.27m, 11.7\textdegree    & 0.03m, 1,74\textdegree                                               & 0.03m, 1.53\textdegree                                                 & 0.02m, 0.9\textdegree   & 0.023m, 0.91\textdegree & 0.26m, - & 0.41m, 15.7\textdegree   & 0.018m, 1.39\textdegree  & 0.26m, 10.92\textdegree & 0.023m, 0.90\textdegree \\ \cline{2-12} 
& heads            & 0.18m, 13.3\textdegree    & 0.05m, 1.98\textdegree                                               & 0.02m, 1.45\textdegree                                                & 0.01m, 0.8\textdegree   & 0.018m, 1.26\textdegree & 0.21m, - & 0.28m, 13.01\textdegree  & 0.016m, 0.99\textdegree  & 0.17m, 12.70\textdegree & 0.014m, 0.82\textdegree \\ \cline{2-12} 
& office           & 0.17m, 5.15\textdegree    & 0.04m, 1.62\textdegree                                               & 0.09m, 3.61\textdegree                                                & 0.03m, 0.7\textdegree   & 0.026m, 0.73\textdegree & 0.36m, - & 0.43m, 7.65\textdegree   & 0.024m, 1.14\textdegree  &0.18m, 5.45\textdegree & 0.025m, 0.69\textdegree \\ \cline{2-12} 
& pumpkin          & 0.22m, 4.02\textdegree    & 0.04m, 1.64\textdegree                                               & 0.08m, 3.10\textdegree                                               & 0.04m, 1.1\textdegree   & 0.039m, 1.09\textdegree & 0.31m, - & 0.49m, 10.63\textdegree  & 0.024m, 1.45\textdegree  & 0.20m, 3.69\textdegree & 0.037m, 1.02\textdegree \\ \cline{2-12} 
& redkitchen       & 0.23m, 4.93\textdegree    & 0.04m, 1.63\textdegree                                              & 0.07m, 3.37\textdegree                                                & 0.04m, 1.1\textdegree   & 0.039m, 1.18\textdegree & 0.26m, - & 0.57m, 8.53\textdegree   & 0.025m, 2.27\textdegree  &0.23m, 4.92\textdegree  & 0.038m, 1.16\textdegree \\ \cline{2-12} 
& stairs           & 0.30m, 12.1\textdegree    & 0.04m, 1.51\textdegree                                               & 0.03m, 2.22\textdegree                                               & 0.09m, 2.6\textdegree   & 0.037m, 1.06\textdegree & 0.14m, - & 0.46m, 14.56\textdegree  & 0.021m,1.08\textdegree  & 0.23m, 11.3\textdegree  & 0.033m, 0.94\textdegree \\ \cline{2-12} 
& \textbf{Average} & 0.207m, 7.78\textdegree    & 0.040m, 1.69\textdegree                                               & 0.051m, 2.46\textdegree                                              & 0.036m, 1.10\textdegree & \textbf{0.029m, 0.98\textdegree} & 0.246m, - & 0.424m, 11.00\textdegree & \textbf{0.022m}, 1.39\textdegree  & 0.190m, 7.47\textdegree & 0.027m, \textbf{0.88}\textdegree \\ 
\Xhline{2\arrayrulewidth}

\multirow{7}{*}{\rotatebox[origin=c]{90}{Cambridge}} 
& GreatCourt       & - & - & -                                                                                                    & 0.40m, 0.2\textdegree    & 0.43m, 0.20\textdegree  & -        & -             & -     &-       &     0.42m,   0.21\textdegree       \\ \cline{2-12} 
& KingsCollege     & 1.07m, 1.89\textdegree  & -                                                 & 0.42m, 0.55\textdegree                                               & 0.18m, 0.3\textdegree   &     0.16m, 0.29\textdegree         & -        & 2.01m, 5.35\textdegree   & - &- &             0.16m, 0.27\textdegree \\ \cline{2-12} 
& OldHospital      & 1.94m, 3.91\textdegree     & -                                              & 0.44m, 1.01\textdegree                                              & 0.20m, 0.3\textdegree   & 0.18m, 0.29\textdegree  & -        & 2.35m, 5.05\textdegree   & - &- &    0.18m, 0.28\textdegree          \\ \cline{2-12} 
& ShopFacade       & 1.49m, 4.22\textdegree    & -                                               & 0.12m, 0.40\textdegree                                               & 0.06m, 0.3\textdegree   & 0.05m, 0.34\textdegree & -        & 1.63m, 6.89\textdegree   & - &- & 0.05m, 0.31\textdegree \\ \cline{2-12} 
& StMarysChurch    & 2.00m, 4.53\textdegree    & -                                               & 0.19m, 0.54\textdegree                                              & 0.13m, 0.4\textdegree   & 0.12m, 0.36\textdegree  & -        & 2.61m, 8.94\textdegree   & - & - &    0.12m, 0.35\textdegree          \\ \cline{2-12} 
& Street           & -     & -                                              & 0.85m, 0.83\textdegree                                              & -            & -            & -        & 3.05m, 5.62\textdegree   & -       & -     &      -        \\ \cline{2-12} 
& \textbf{Average} \tnote{1}
& 1.63m, 3.64\textdegree        & -                                           & 0.29m, 0.63\textdegree                                   & 0.14m, 0.33\textdegree  &  \textbf{0.13m, 0.32\textdegree}            &   -       & 2.15m, 6.56\textdegree   & -  & -&     \textbf{0.13m, 0.30\textdegree}        \\ 
\Xhline{2\arrayrulewidth}
\multicolumn{2}{c?}{DeepLoc}   &  - & - & \textbf{0.010m, 0.04\textdegree} & - & 0.083m, 0.45\textdegree & - & - & 0.320m, 1.48\textdegree & - & \textbf{0.065m, 0.43\textdegree}      \\
\Xhline{2\arrayrulewidth}                                                                             
\end{tabular}
}
\begin{tablenotes}
\scriptsize
	\item[1] The average does not include errors of \textit{GreatCourt} and \textit{Street} as some methods do not report results of the two scenes.
\end{tablenotes}
}
\end{threeparttable}
\caption{The median translation and rotation errors of different relocalization methods. Best results are in bold.}
\vspace{-1em}
\label{table:pose_error}
\end{table*}

Normally, the innovation variable $\mathbf{e}_{(i)} \in \mathbb{R}^3$ follows the Gaussian distribution $\mathcal{N}(\mathbf{0}, \mathbf{S}_{(i)})$ as shown by Eq.~\ref{eq:bivariate}, where  $\mathbf{S}_{(i)} = ({{v}_{(i)}}^2 + {{r}_{(i)} }^2) \mathbb{I}_3$.
Then, $\textrm{NIS}  = \mathbf{e}_{(i)}^T \mathbf{S}_{(i)}^{-1} \mathbf{e}_{(i)}$ is supposed to follow the Chi-squared distribution with three degrees of freedom, denoted as $\chi^2(3)$.
It is thus reasonable to see a pixel state as an outlier if its NIS value locates outside the acceptance region of $\chi^2(3)$.
As illustrated in Fig.~\ref{fig:chi-square}, we use the critical value of $0.05$ in the NIS test, which means we have at least $95\%$ statistical evidence to regard one pixel state as negative.
The uncertainties of the pixels failing the test, \eg $\sigma_{(i)}$, are reset to be infinitely large so that they will have no effect in later steps.

\begin{table}[]
\centering
\resizebox{0.6\linewidth}{!}
{
\begin{tabular}{c|c|c|c}
\Xhline{2\arrayrulewidth}                                                                             
\multicolumn{3}{c|}{One-shot} & Temporal \\ \hline
DSAC++\cite{brachmann2018learning}    & ESAC \cite{esac}   & SCoordNet  & KFNet    \\ \hline \hline
96.8\%     & 97.8\%   & 98.9\%       & \textbf{99.2\%}     \\ 
\Xhline{2\arrayrulewidth}                                                                             
\end{tabular}
}
\caption{The 5cm-5deg accuracy of one-shot and temporal relocalization methods on 12scenes \cite{valentin2016learning}.}
\vspace{-1em}
\label{table:pose_accuracy}
\end{table}

\section{Experiments} \label{sec:experiments}

\subsection{Experiment Settings}

\smallskip\noindent\textbf{Datasets.}
Following previous works \cite{kendall2015posenet,brachmann2017dsac,brachmann2018learning,radwan2018vlocnet++}, we use two indoor datasets - \textit{7scenes} \cite{shotton2013scene} and \textit{12scenes} \cite{valentin2016learning}, and two outdoor datasets - \textit{DeepLoc} \cite{radwan2018vlocnet++} and \textit{Cambridge} \cite{kendall2015posenet} for evaluation.  
\iffalse
Through the four datasets, the scale grows larger and the image acquisition gets sparser in time domain.
Besides, the motion blur, illumination changes, scene aliasing conditions and the dynamic environment add to the variation of the datasets.
\fi
Each scene has been split into different strides of sequences for training and testing. 

\smallskip\noindent\textbf{Data processing.}
Images are downsized to $640\times480$ for \textit{7scenes} and \textit{12scenes}, $848\times480$ for \textit{DeepLoc} and \textit{Cambridge}.
The groundtruth scene coordinates of \textit{7scenes} and \textit{12scenes} are computed based on given camera poses and depth maps, whereas those of \textit{DeepLoc} and \textit{Cambridge} are rendered from surfaces reconstructed with training images.
\iffalse
We decorrelate the point coordinates of every scene to give zero means and correlations.
During training, the images are randomly rescaled by $0.8 \sim 1.2$ times and rotated by $\pm \pi/6$. The images colors are augmented by adjusting the contrast by $0.8 \sim 1.2$ times and shifting the brightness by $\pm 20$.
\fi

\begin{figure*}[t]
\begin{center}
\includegraphics[width=0.98\linewidth]{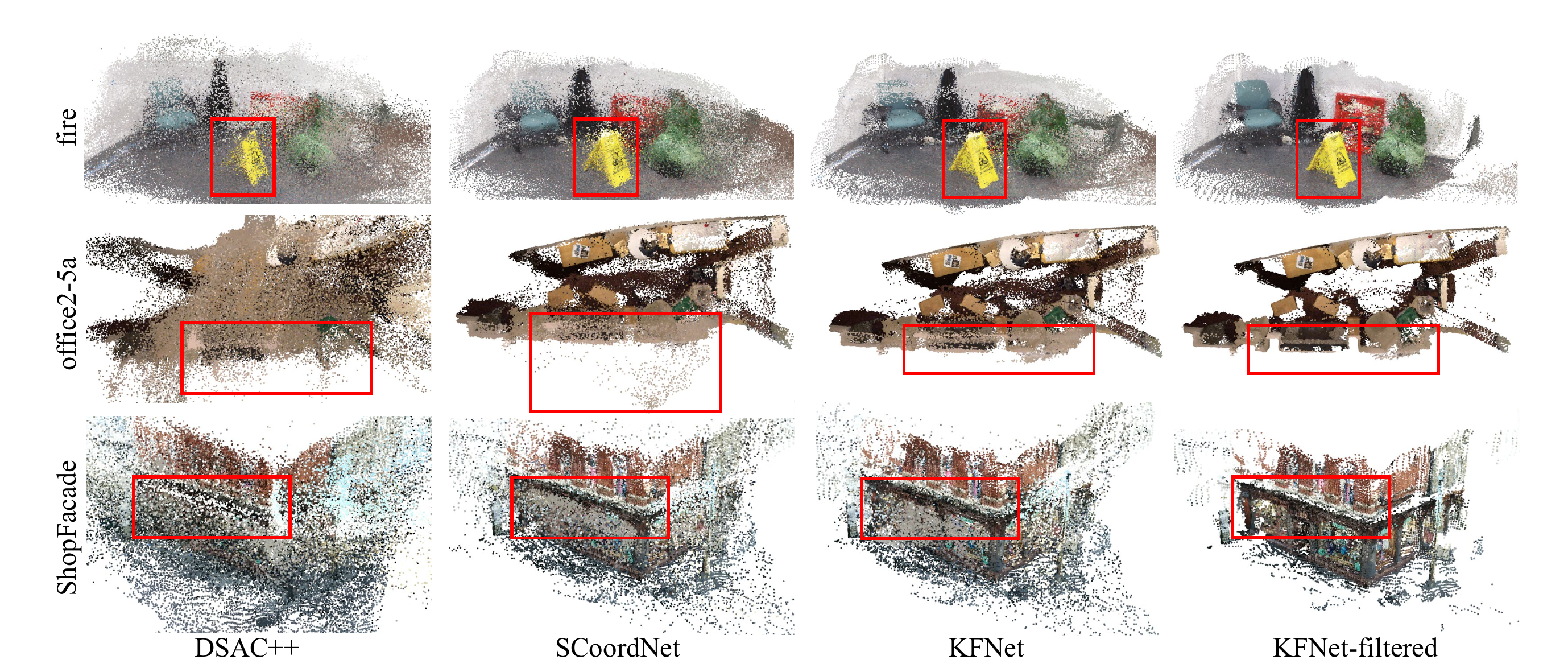}
\end{center}
  \caption{The point clouds predicted by different relocalization methods. Our SCoordNet and KFNet increasingly suppress the noise as highlighted by the red boxes and produce much neater point clouds than the state-of-the-art DSAC++ \cite{brachmann2018learning}. The KFNet-filtered panel filters out the points of KFNet of which the uncertainties are too large and gives rather clean and accurate mapping results.}
  \vspace{-1em}
\label{fig:point_clouds}
\end{figure*}

\smallskip\noindent\textbf{Training.}
Our best practice chooses the parameter setting as  $\tau_1=0.2$, $\tau_2=0.2$, $\tau_3=0.6$.
The ADAM optimizer \cite{kingma2014adam} is used with $\beta_1=0.9$ and $\beta_2=0.999$.
We use an initial learning rate of $\gamma=0.0001$ and then drop it with exponential decay.
The training procedure has 3 stages.
First, we train SCoordNet for each scene with the likelihood loss $\mathcal{L}_{likelihood} $ (Eq.~\ref{eq:likelihood_loss}).
The iteration number is set to be proportional to the surface area of each scene and the learning rate drops from $\gamma$ to $\gamma/2^5$.
In particular, we use SCoordNet as the one-shot version of the proposed approach.
Second, OFlowNet is trained using all the scenes for each dataset with the prior loss $\mathcal{L}_{prior} $ (Eq.~\ref{eq:prior_loss}).
It also experiences the learning rate decaying from $\gamma$ to $\gamma/2^5$.
Each batch is composed of two consecutive frames.
The window size of OFlowNet in the original images is set to 64, 128, 192 and 256 for the four datasets mentioned above, respectively, due to the increasing ego-motion through them.
Third, we fine-tune all the parameters of KFNet jointly by optimizing the full loss $\mathcal{L}_{full}$ (Eq.~\ref{eq:full_loss}) with a learning rate going from $\gamma/2^4$ to $\gamma/2^5$.
Each batch in the third stage contains four consecutive frames.
\iffalse
All the experiments run on a machine with a 8-core Intel i7-4770K, a 32GB memory and a
NVIDIA GTX 1080 Ti graphics card.
\fi

\subsection{Results}

\subsubsection{The Relocalization Accuracy} \label{sec:localization_accuracy}

Following \cite{brachmann2017dsac, brachmann2018learning,clark2017vidloc,valada2018deep}, we use two accuracy metrics:
(1) the median rotation and translation error of poses (see Table~\ref{table:pose_error});
(2) the 5cm-5deg accuracy (see Table~\ref{table:pose_accuracy}), \ie, the mean percentage of the poses with translation and rotation errors less than 5 cm and 5\textdegree, respectively.
The uncertainty threshold $\lambda$ (Sec.~\ref{sec:formulation}) is set to 5 cm for \textit{7scenes} and \textit{12scenes} and 50 cm for \textit{DeepLoc} and \textit{Cambridge}.

\smallskip\noindent\textbf{One-shot relocalization.}
Our SCoordNet achieves the lowest pose errors on \textit{7scenes} and \textit{Cambridge}, and the highest 5cm-5deg accuracy on \textit{12scenes} among the one-shot methods, surpassing CamNet \cite{Ding_2019_ICCV} and MapNet \cite{Brahmbhatt_2018_CVPR} which are the state-of-the-art relative and absolute pose regression methods, respectively. Particularly, SCoordNet outperforms the state-of-the-art structure-based methods DSAC++ \cite{brachmann2018learning} and ESAC \cite{esac}, yet with fewer parameters ($24$M vs. $210$M vs. $28$M, respectively). 
The advantage of SCoordNet should be mainly attributed to the uncertainty modeling, as we will analyze in Appendix~\ref{sec:uncertainty_modeling}.
It also surpasses Active Search (AS) \cite{sattler2017efficient} on \textit{7scenes} and \textit{Cambridge}, but underperforms AS on \textit{DeepLoc}. 
We find that, in the experiments of AS on \textit{DeepLoc} \cite{sattler2019understanding}, AS is tested on a SfM model built with both training and test images. This may explain why AS is surprisingly more accurate on \textit{DeepLoc} than on other datasets, since the 2D-3D matches between test images and SfM tracks have been established and their geometry has been optimized during the SfM reconstruction.

\smallskip\noindent\textbf{Temporal relocalization.}
Our KFNet improves over SCoordNet on all the datasets as shown in Tables~\ref{table:pose_error}~\&~\ref{table:pose_accuracy}.
The improvement on \textit{Cambridge} is marginal as the images are over-sampled from videos sparsely. The too large motions between frames make it hard to model the temporal correlations.
KFNet obtains much lower pose errors than other temporal methods, except that it has a larger translation error than VLocNet++ \cite{radwan2018vlocnet++} on \textit{7scenes}.
However, the performance of VLocNet++ is inconsistent across different datasets. On \textit{DeepLoc}, the dataset collected by the authors of VLocNet++, VLocNet++ has a much larger pose error than KFNet, even though it also integrates semantic segmentation into learning.
The inconsistency is also observed in \cite{sattler2019understanding}, which shows that VLocNet++ cannot substaintially exceed the accuracy of retrieval based methods \cite{torii201524,arandjelovic2016netvlad}.

\begin{table}[]
\centering
\resizebox{1\linewidth}{!}
{
\begin{tabular}{c|c|c|c|c|c|c|c|c}
\Xhline{2\arrayrulewidth}
          & \multicolumn{2}{c|}{7scenes} 
          & \multicolumn{2}{c|}{12scenes}
          & \multicolumn{2}{c|}{DeepLoc} 
          & \multicolumn{2}{c}{Cambridge} \\ \hline 
          & mean         & stddev        & mean         & stddev         & mean          & stddev  & mean          & stddev         \\ \hline\hline
DSAC++ \cite{brachmann2018learning}    & 28.8         & 33.1          & 28.8         & 47.1        &  -  &  -   & 467.3         & 883.7          \\ \hline
SCoordNet & 16.8         & 23.3          & 9.8          & 20.0     &  883.0 &  1520.8    & 272.7         & 497.6          \\ \hline
KFNet     & \textbf{15.3}         & \textbf{21.7}          & \textbf{7.3}          & \textbf{13.7}      &  \textbf{200.79} & \textbf{398.8}    & \textbf{241.5}         & \textbf{441.7}          \\ 
\Xhline{2\arrayrulewidth}
\end{tabular}
}
\caption{The mean and standard deviation of predicted scene coordinate errors in centimeters.}
\vspace{-1em}
\label{table:scene_coord_error}
\end{table}

\subsubsection{The Mapping Accuracy}
Relocalization methods based on SCoRe \cite{shotton2013scene,brachmann2018learning} can create a mapping result for each view by predicting per-pixel scene coordinates.
Hence, relocalization and mapping can be seen as dual problems, as one can be easily resolved once the other is known.
Here, we would like to evaluate the mapping accuracy with the mean and the standard deviation (stddev) of scene coordinate errors of the test images.

As shown in Table~\ref{table:scene_coord_error}, the mapping accuracy is in accordance with the relocalization accuracy reported in Sec.~\ref{sec:localization_accuracy}.
SCoordNet reduces the mean and stddev values greatly compared against DSAC++, and KFNet further reduces the mean error over SCoordNet by $8.9\%$, $25.5\%$, $77.3\%$ and $11.4\%$ on the four datasets, respectively.
The improvements are also reflected in the predicted point clouds, as visualized in Fig.~\ref{fig:point_clouds}. 
SCoordNet and KFNet predict less noisy scene points with better temporal consistency compared with DSAC++.
Additionally, we filter out the points of KFNet with uncertainties greater than $\lambda$ as displayed in the KFNet-filtered panel of Fig.~\ref{fig:point_clouds}, which helps to give much neater and more accurate 3D point clouds.

\subsubsection{Motion Blur Experiments}

\begin{figure}[]
\begin{center}
\includegraphics[width=1\linewidth]{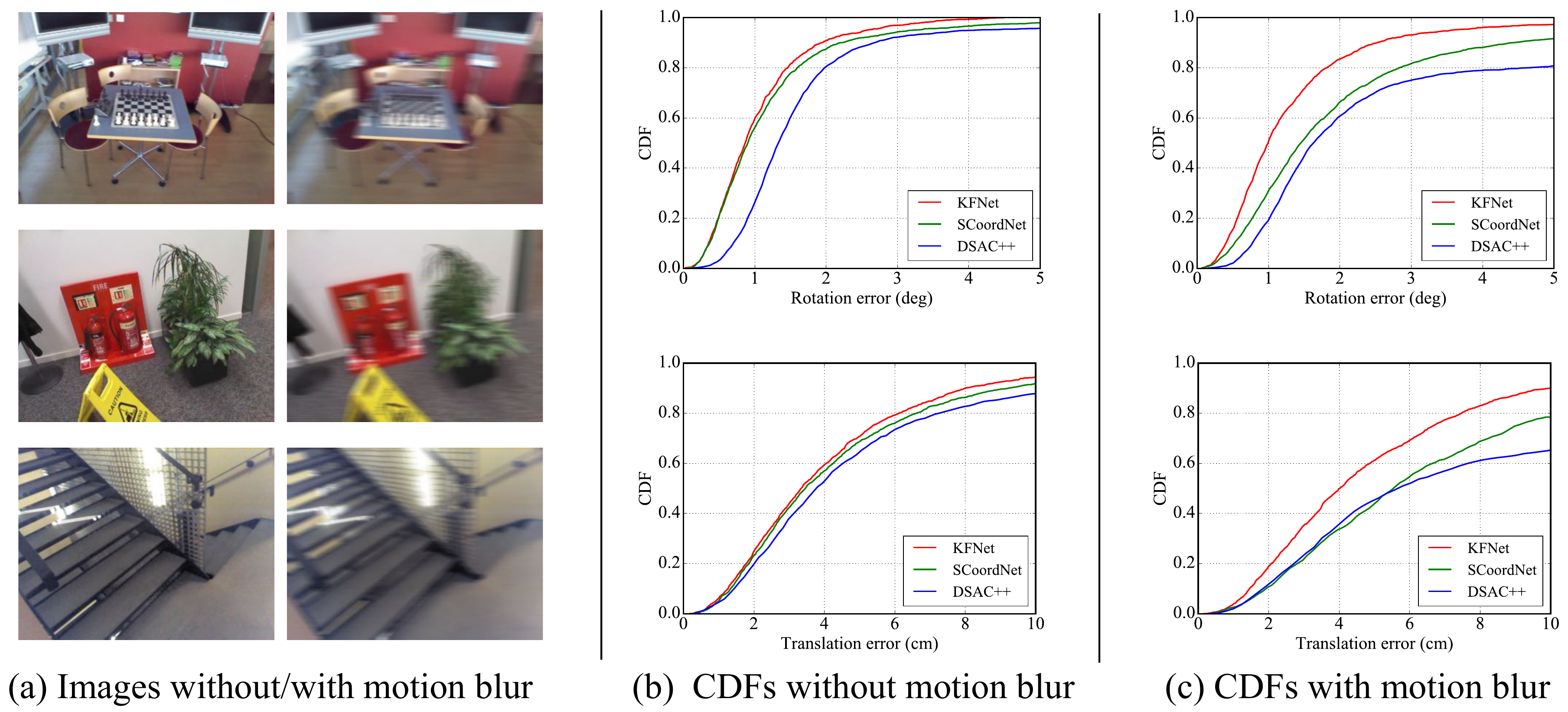}
\end{center}
  \caption{(a) Artificial motion blur images. (b) \& (c) The cumulative distribution functions (CDFs) of pose errors before and after motion blur is applied. }
  \vspace{-1em}
\label{fig:motion_blur2}
\end{figure}

Although, in terms of the \textit{mean scene coordinate error} in Table.~\ref{table:scene_coord_error}, SCoordNet outperforms DSAC++ by over $41.6\%$ and KFNet further improves SCoordNet by a range from $8.9\%$ to $77.3\%$, the improvements in terms of the \textit{median pose error} in Table~\ref{table:pose_error} are not as significant.
The main reason is that the RANSAC-based PnP solver diminishes the benefits brought by the scene coordinate improvements, since only a small subset of accurate scene coordinates selected by RANSAC matters in the pose accuracy.
Therefore, to highlight the advantage of KFNet, we conduct more challenging  experiments over motion blur images which are quite common in real scenarios.
For the test image sequences of \textit{7scenes}, we apply a motion blur filter with a kernel size of 30 pixels for every 10 images as shown in Fig.~\ref{fig:motion_blur2}(a).
In Fig.~\ref{fig:motion_blur2}(b)\&(c), we plot the cumulative distribution functions of the pose errors before and after applying motion blur.
Thanks to the uncertainty reasoning, SCoordNet generally attains smaller pose errors than DSAC++ whether motion blur is present.
While SCoordNet and DSAC++ show a performance drop after motion blur is applied, KFNet maintain the pose accuracy as shown in Fig.~\ref{fig:motion_blur2}(b)\&(c), 
leading to a more notable margin between KFNet and SCoordNet and demonstrating the benefit of the temporal modelling used by KFNet.

\subsection{Ablation studies}

\begin{table}[t]
\centering
\resizebox{1\linewidth}{!}
{
\begin{tabular}{c|c|c|c|c}
\Xhline{2\arrayrulewidth}
One-shot     & \multicolumn{4}{c}{Temporal}                             \\ \hline
SCoordNet    & ConvLSTM  \cite{xingjian2015convolutional} 
   & TPooler  \cite{pfister2015flowing} 
       & SWeight  \cite{Zhu_2017_ICCV}      & KFNet        \\ \hline \hline
0.029m, 0.98\textdegree & 0.040m, 1.12\textdegree & 0.029m, 0.94\textdegree & 0.029m, 0.95\textdegree & \textbf{0.027m, 0.88\textdegree } \\ 
\Xhline{2\arrayrulewidth}
\end{tabular}
}
\caption{The median pose errors produced by different temporal aggregation methods on \textit{7scenes}. Our KFNet achieves better pose accuracy than other temporal aggregation strategies.}
\vspace{-1em}
\label{table:temporal}
\end{table}

\paragraph{Evaluation of Temporal Aggregation.}
This section studies the efficacy of our Kalman filter based framework in comparison with other popular temporal aggregation strategies including ConvLSTM \cite{xingjian2015convolutional,Lai_2018_ECCV}, temporal pooler (TPooler) \cite{pfister2015flowing} and similarity weighting (SWeight) \cite{Zhu_2017_ICCV,Zhu_2018_CVPR}.
KFNet is more related to TPooler and SWeight which also use the flow-guided warping yet within an n-frame neighborhood.
For equitable comparison, the same feature network and probabilistic losses as KFNet are applied to all.
We use a kernel size of $8$ for ConvLSTM to ensure a window size of $64$ in images.
The same OFlowNet structure and a $3$-frame neighborhood are used for TPooler and SWeight for flow-guided warping.

Table~\ref{table:temporal} shows the comparative results on \textit{7scenes}.
ConvLSTM largely underperforms SCoordNet and other aggregation methods in pose accuracy, which manifests the necessity of explicitly determining the pixel associations between frames instead of implicit modeling.
Although the flow-guided warping is employed, TPooler and SWeight only achieve marginal improvements over SCoordNet compared with KFNet, which justifies the advantage of the Kalman filtering system.
Compared with TPooler and SWeight, the Kalman filter behaves as a more disciplined and non-heuristic approach to temporal aggregation that ensures an optimal solution of the linear Gaussian state-space model \cite{fruhwirth1995bayesian} defined in Sec.~\ref{sec:formulation}.

\begin{figure}[]
\begin{center}
\includegraphics[width=1\linewidth]{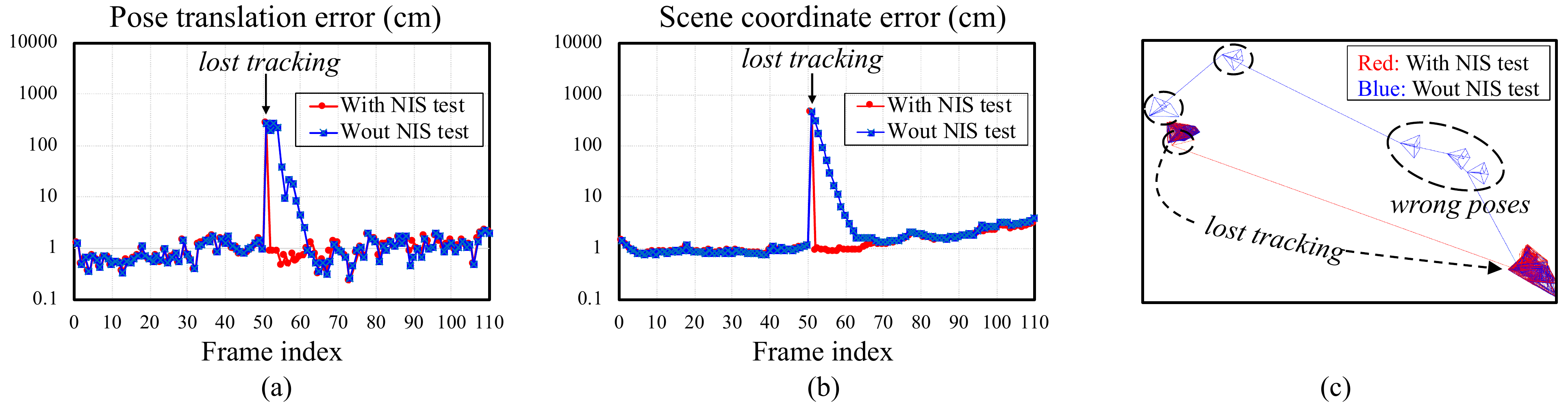}
\end{center}
  \caption{(a)~\&~(b) With NIS testing \cite{bar2004estimation}, the errors of poses and scene coordinates quickly revert to normal after the lost tracking. (c) The poses of a sample sequence show that, without NIS testing, the lost tracking adversely affects the pose accuracy of the subsequent frames.}
  \vspace{-1em}
\label{fig:consistency_checking}
\end{figure}

  \vspace{-0.5em}
\paragraph{Evaluation of Consistency Examination}
  \vspace{-0.2em}
Here, we explore the functionality of the consistency examination which uses NIS testing \cite{bar2004estimation} (see Sec.~\ref{sec:consistency}). 
Due to the infrequent occurrence of extreme outlier predictions among the well-built relocalization datasets,
we simulate the tracking lost situations by
trimming a sub-sequence off each testing sequence of \textit{7scenes} and \textit{12scenes}.
Let $\mathbf{I}_p$ and $\mathbf{I}_q$ denote the last frame before and the first frame after the trimming.
The discontinuous motion from $\mathbf{I}_p$ to $\mathbf{I}_q$ would cause outlier scene coordinate predictions for $\mathbf{I}_q$ by KFNet.
Fig.~\ref{fig:consistency_checking} plots the mean pose and scene coordinate errors of frames around $\mathbf{I}_q$ and visualizes the poses of a sample trimmed sequence.
With the NIS test, the errors revert to a normal level promptly right after $\mathbf{I}_q$, whereas without the NIS test, the accuracy of poses after $\mathbf{I}_q$ is affected adversely.
NIS testing stops the propagation of the outlier predictions of $\mathbf{I}_q$ into later steps by giving them infinitely large uncertainties, so that $\mathbf{I}_{q+1}$ will leave out the prior from $\mathbf{I}_q$ and reinitialize itself with the predictions of the measurement system.

\section{Conclusion}

This work addresses the temporal camera relocalization problem by proposing a recurrent network named KFNet.
It extends the scene coordinate regression problem to the time domain for online pose determination.
The architecture and the loss definition of KFNet are based on the Kalman filter, which allows a disciplined manner of aggregating the pixel-level predictions through time.
The proposed approach yields the top accuracy among the state-of-the-art relocalization methods over multiple benchmarks. 
Although KFNet is only validated on the camera relocalization task, the immediate application alongside other tasks like video processing \cite{Kim_2018_ECCV, Lai_2018_ECCV} and segmentation \cite{valipour2017recurrent,Nilsson_2018_CVPR} , object tracking \cite{Lu_2017_ICCV,Zhu_2018_CVPR} would be anticipated.

\clearpage

\appendix

\noindent{\LARGE \textbf{Appendices}}
\section{Full Network Architecture} \label{sec:architecture}

As a supplement to the main paper, we detail the parameters of the layers of SCoordNet and OFlowNet used for training \textit{7scenes} in Table~\ref{table:network} at the end of the appendix.

\section{Supplementary Derivation of the Bayesian Formulation} \label{sec:bayesian_derivation}

This section supplements the derivation of the distributions \ref{eq:bivariate} \& \ref{eq:posterior_derived} in the main paper.

Let us denote the bivariate Gaussian distribution of the latent state $\boldsymbol{\theta}_t$ and the innovation $\mathbf{e}_t$ conditional on $\mathcal{I}_{t-1}$ as
\begin{equation}   \label{eq:bivariate_general}
\left[  \left. \begin{pmatrix} \boldsymbol{\theta}_t \\ \mathbf{e}_t \end{pmatrix} \right\vert \mathcal{I}_{t-1} \right] \sim \mathcal{N} \left[ \begin{pmatrix} \boldsymbol{\mu}_1 \\ \boldsymbol{\mu}_2 \end{pmatrix} , \begin{pmatrix}  \boldsymbol{\Sigma}_{11} & \boldsymbol{\Sigma}_{12} \\ \boldsymbol{\Sigma}_{21} & \boldsymbol{\Sigma}_{22} \end{pmatrix}  \right],
\end{equation}
where $ \boldsymbol{\Sigma}_{12} =  {\boldsymbol{\Sigma}_{21}}^T$.
Based on the multivariate statistics theorems \cite{anderson1984introduction}, the conditional distribution of $\boldsymbol{\theta}_t$ given $\mathbf{e}_t$ is expressed as $(\boldsymbol{\theta}_t  \vert \mathbf{e}_t, \mathcal{I}_{t-1})  \sim$
\begin{equation} \label{eq:latent_conditional}
 \mathcal{N}(\boldsymbol{\mu}_1 + \boldsymbol{\Sigma}_{12} \boldsymbol{\Sigma}_{22}^{-1} (\mathbf{e}_t  - \boldsymbol{\mu}_2),  \boldsymbol{\Sigma}_{11} - \boldsymbol{\Sigma}_{12}\boldsymbol{\Sigma}_{22}^{-1}\boldsymbol{\Sigma}_{21}),
\end{equation}
and similarly, $(\mathbf{e}_t  \vert \boldsymbol{\theta}_t, \mathcal{I}_{t-1}) \sim$
\begin{equation} \label{eq:innovation_conditional}
\mathcal{N}(\boldsymbol{\mu}_2 + \boldsymbol{\Sigma}_{21} \boldsymbol{\Sigma}_{11}^{-1} (\boldsymbol{\theta}_t - \boldsymbol{\mu}_1),  \boldsymbol{\Sigma}_{22} - \boldsymbol{\Sigma}_{21}\boldsymbol{\Sigma}_{11}^{-1}\boldsymbol{\Sigma}_{12}).
\end{equation}
Conversely, if Eq.~\ref{eq:latent_conditional} holds and $(\boldsymbol{\theta}_t \vert \mathcal{I}_{t-1}) \sim \mathcal{N}(\boldsymbol{\mu}_1, \boldsymbol{\Sigma}_{11})$, Eq.~\ref{eq:bivariate_general} will also hold according to \cite{anderson1984introduction}.
Since we have had $(\boldsymbol{\theta}_t | \mathcal{I}_{t-1} ) \sim   \mathcal{N}(\hat{\boldsymbol{\theta}}_t^{-}, \mathbf{R}_t)$ in Eq.~\ref{eq:distribution_transition} of the main paper, we can note that
\begin{equation} \label{eq:variables1}
\boldsymbol{\mu}_1 = \hat{\boldsymbol{\theta}}_t^{-}, \;\; \text{and} \;\; \boldsymbol{\Sigma}_{11} = \mathbf{R}_t.
\end{equation}
Recalling Eq.~\ref{eq:distribution_measurement} of the main paper, we already have
\begin{equation}  \label{eq:innovation_conditional_actual}
(\mathbf{e}_t | \boldsymbol{\theta}_t ,\mathcal{I}_{t-1}) \sim \mathcal{N}(\boldsymbol{\theta}_t -  \hat{\boldsymbol{\theta}}_t^{-}, \mathbf{V}_t).
\end{equation}
Equalizing Eq.~\ref{eq:innovation_conditional} and Eq.~\ref{eq:innovation_conditional_actual}, we have
\begin{equation} \label{eq:variables2}
\begin{split}
\boldsymbol{\mu}_2 &= \mathbf{0}, \\
\boldsymbol{\Sigma}_{12} &= \boldsymbol{\Sigma}_{21} = \mathbf{R}_t, \\
\boldsymbol{\Sigma}_{22} &= \mathbf{V}_t + \mathbf{R}_t.
\end{split}
\end{equation}
Substituting the variables of Eqs.~\ref{eq:variables1}~\&~\ref{eq:variables2} into Eqs.~\ref{eq:bivariate_general}~\&~\ref{eq:latent_conditional}, we have reached the distributions \ref{eq:bivariate} \& \ref{eq:posterior_derived} in the main paper.

\begin{figure}[]
\begin{center}
\includegraphics[width=1\linewidth]{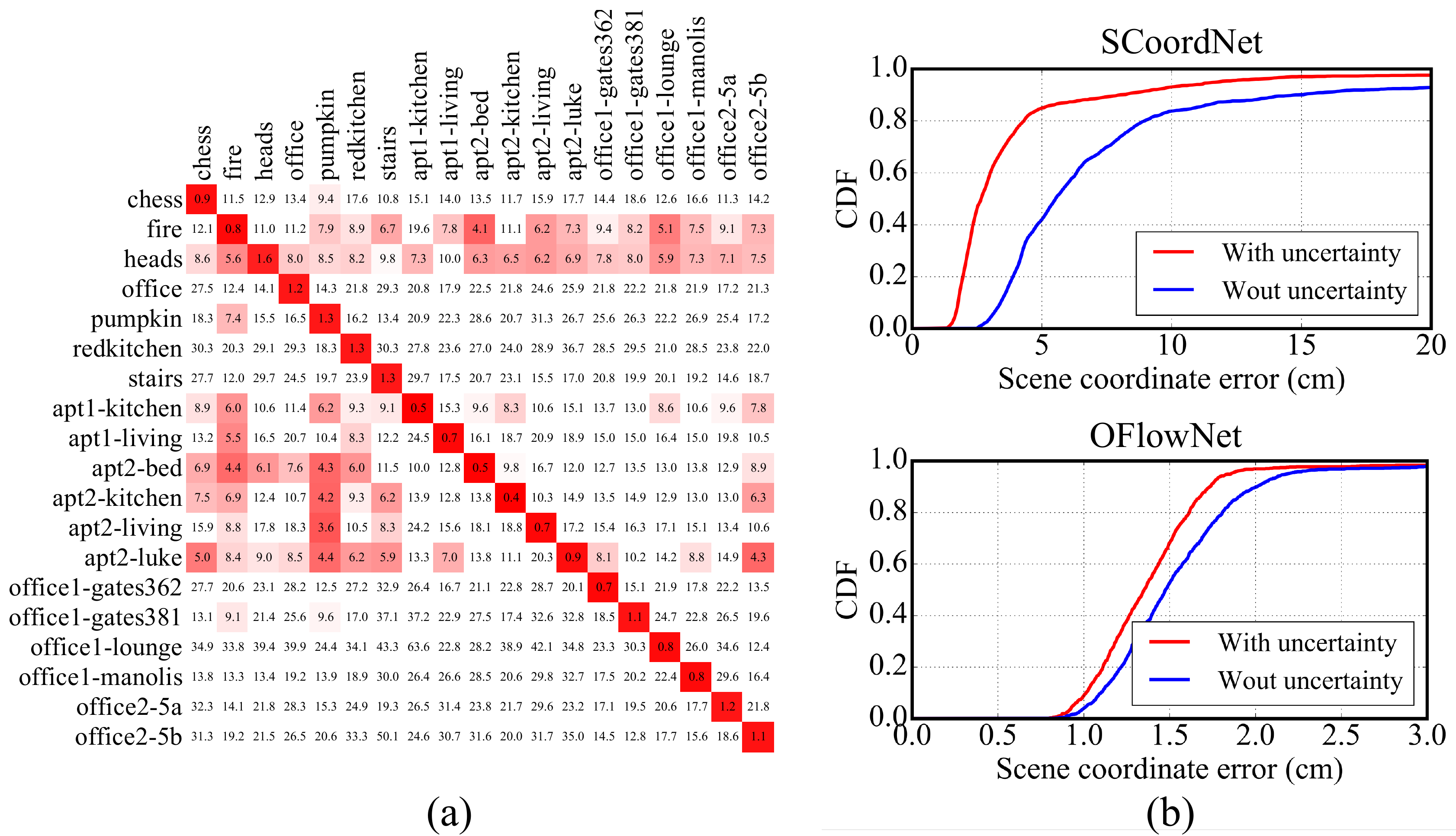}
\end{center}
  \caption{(a) The confusion matrix of 19 scenes given by our uncertainty predictions. The redder a block (i, j), the more likely it is that the images of the j-th scene belong to the i-th scene. (b) The CDFs of scene coordinate errors given by SCoordNet and OFlowNet with or without uncertainty modeling. }
  \vspace{-1em}
\label{fig:uncertainty_ablation_paper}
\end{figure}

\section{Ablation Study on the Uncertainty Modeling} \label{sec:uncertainty_modeling}
The uncertainty modeling, which helps to quantify the measurement and process noise,
is an indispensable component of KFNet. 
In this section, we conduct ablation studies on it.

First, we run the trained KFNet of each scene from \textit{7scenes} and \textit{12scenes} over the test images of each scene exhaustively and visualize the median uncertainties as the confusion matrix in Fig.~\ref{fig:uncertainty_ablation_paper}(a).
The uncertainties between the same scene in the main diagonal are much lower than those between different scenes.
It indicates that meaningful uncertainties are learned which can be used for scene recognition.
Second, we qualitatively compare SCoordNet and OFlowNet against their counterparts which are trained with L2 loss without uncertainty modeling.
The cumulative distribution functions (CDFs) of scene coordinate errors tested on \textit{7scenes} and \textit{12scenes} are shown in Fig.~\ref{fig:uncertainty_ablation_paper}(b).
The uncertainty modeling leads to more accurate predictions for both SCoordNet and OFlowNet.
We attribute the improvements to the fact that the uncertainties apply auto-weighting to the loss term of each pixel as in Eqs.~\ref{eq:likelihood_loss} \& \ref{eq:prior_loss} of the main paper, which prevents the learning from getting stuck in the hard or infeasible examples like the boundary pixels for SCoordNet and the occluded pixels for OFlowNet (see Fig.~\ref{fig:uncertainty} of the main paper).

\begin{table}[t]
\centering
\resizebox{1\linewidth}{!}
{
\begin{tabular}{c|c|c|c|c|c|c|c}
\Xhline{2\arrayrulewidth}
\multirow{2}{*}{\begin{tabular}[c]{@{}c@{}} {Downsample} \\ {Rate} \end{tabular}} & \multirow{2}{*}{\begin{tabular}[c]{@{}c@{}} {Receptive}\\ {field}\end{tabular}} & \multicolumn{6}{c}{ {Layers (kernel, stride)} }            \\ \cline{3-8}  
                                                                           &                                                                            & L7   & L8   & L9   & L10  & L11  & L12  \\ \hline\hline
8                                                                          & 29                                                                         & 1, 2 & 1, 1 & 1, 1 & 1, 1 & 1, 1 & 1, 1 \\ \hline
8                                                                          & 45                                                                         & 3, 2 & 1, 1 & 1, 1 & 1, 1 & 1, 1 & 1, 1 \\ \hline
8                                                                          & 61                                                                         & 3, 2 & 3, 1 & 1, 1 & 1, 1 & 1, 1 & 1, 1 \\ \hline
8                                                                          & 93                                                                         & 3, 2 & 3, 1 & 3, 1 & 3, 1 & 1, 1 & 1, 1 \\ \hline
8                                                                          & 125                                                                        & 3, 2 & 3, 1 & 3, 1 & 3, 1 & 3, 1 & 3, 1 \\ \hline
8                                                                          & 157                                                                        & 3, 2 & 3, 1 & 5, 1 & 5, 1 & 3, 1 & 3, 1 \\ \hline
8                                                                          & 189                                                                        & 3, 2 & 3, 1 & 5, 1 & 5, 1 & 5, 1 & 5, 1 \\ \hline
8                                                                          & 221                                                                        & 3, 2 & 3, 1 & 7, 1 & 7, 1 & 5, 1 & 5, 1 \\ \Xhline{2\arrayrulewidth}
4                                                                          & 93                                                                         & 3, 1 & 3, 1 & 5, 1 & 5, 1 & 3, 1 & 3, 1 \\ \hline
8                                                                          & 93                                                                         & 3, 2 & 3, 1 & 3, 1 & 3, 1 & 1, 1 & 1, 1 \\ \hline
16                                                                         & 93                                                                         & 3, 2 & 3, 1 & 3, 2 & 1, 1 & 1, 1 & 1, 1 \\ \hline
32                                                                         & 93                                                                         & 3, 2 & 3, 1 & 3, 2 & 1, 1 & 1, 2 & 1, 1 \\ \Xhline{2\arrayrulewidth}
\end{tabular}
}
\caption{The parameters of 7-th to 12-th layers of SCoordNet \wrt different downsample rates and receptive fields. The number before comma is kernel size, while the one after comma is stride.}
\label{table:params}
\end{table}

\begin{table}[t]
\centering
\resizebox{1\linewidth}{!}
{
\begin{tabular}{c|c|c|c|c}
\Xhline{2\arrayrulewidth}
\multirow{2}{*}{{\begin{tabular}[c]{@{}c@{}}Receptive \\ field\end{tabular}}} & \multicolumn{2}{c|}{{Relocalization accuracy}} & \multicolumn{2}{c}{{Mapping accuracy}} \\ \cline{2-5}
& pose error           & pose accuracy         & mean              & stddev               \\ \hline \hline
29                       &  0.025m, 0.87\textdegree       &  87.9\%                         &  29.6cm                    &   32.3                   \\ \hline
45                       & 0.023m, 0.88\textdegree         & 93.4\%                         & 24.4cm                    & 29.2                 \\ \hline
61                       & \textbf{0.023m, 0.84\textdegree}         & \textbf{94.0\%}                           & 17.3cm                    & 23.1                 \\ \hline
93                       & 0.024m, 0.91\textdegree         & 92.9\%                           & 11.5cm                    & 16.4                 \\ \hline
125                     & 0.026m, 0.95\textdegree         & 88.3\%                           & 11.7cm                    & 16.1                 \\ \hline
157                      & 0.026m, 0.97\textdegree         & 86.6\%                  & 10.3cm                    & 15.0                 \\ \hline
189                      & 0.030m, 1.07\textdegree         & 81.0\%                           & 10.3cm                    & 13.9                 \\ \hline
221                     & 0.031m, 1.22\textdegree & 71.8\%                           & \textbf{9.5cm}                     & \textbf{12.9}                 \\ 
\Xhline{2\arrayrulewidth}
\end{tabular}
}
\caption{The performance of SCoordNet \wrt the receptive field. The pose accuracy means the percentage of poses with rotation and translation errors less than $5$\textdegree and $5$cm, respectively.}
\label{table:receptive_field_result}
\end{table}

\section{Ablation Study on the Receptive Field} \label{sec:receptive_field}

\begin{figure}[t]
\begin{center}
\includegraphics[width=1\linewidth]{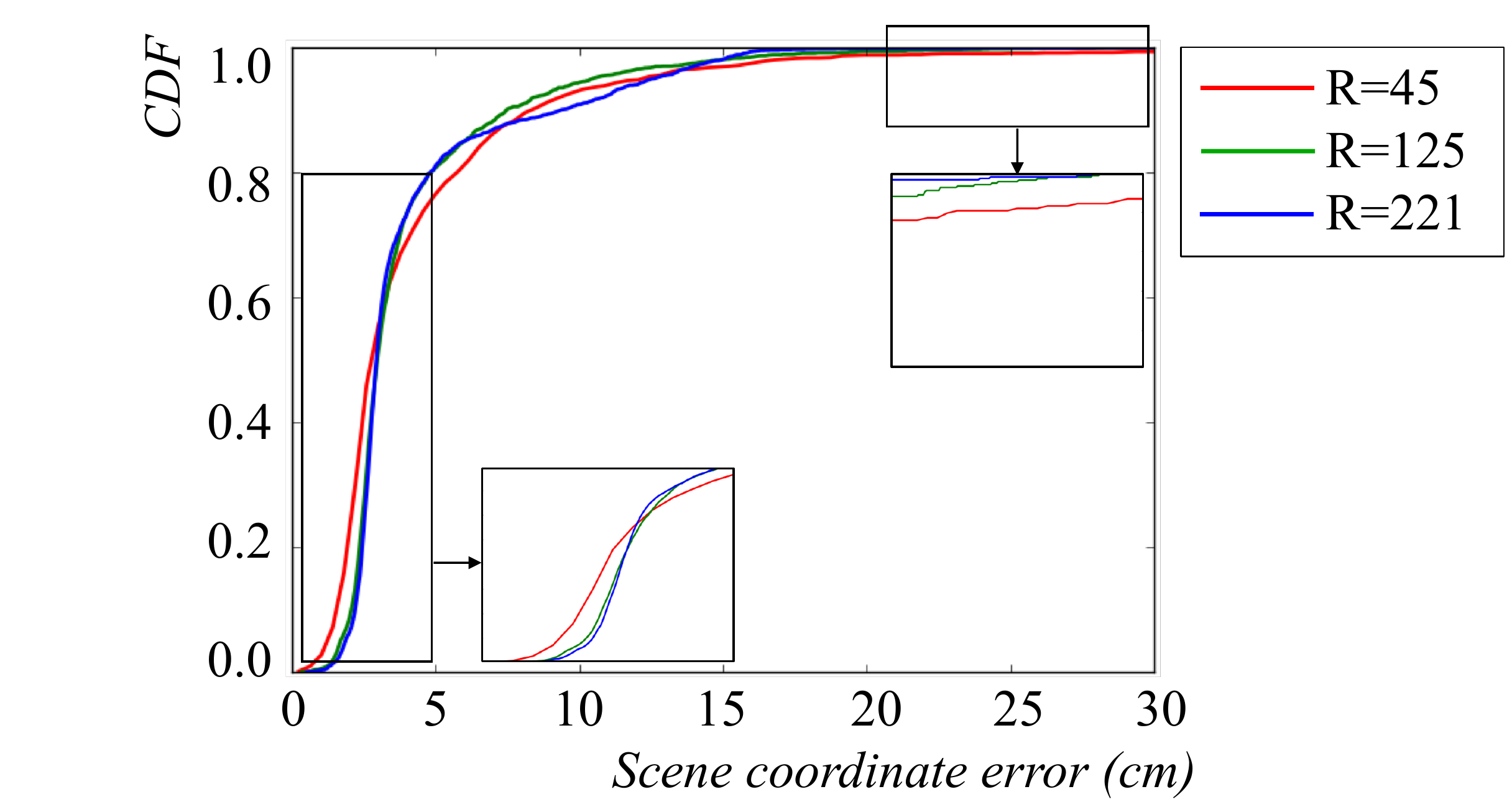}
\end{center}
  \caption{The cumulative distribution function of scene coordinate errors \wrt different receptive field $R$. A smaller $R$ generally has a denser distribution of errors smaller than $2$cm as well as larger than $20$cm. The more predictions with errors smaller than $2$cm contribute to the accuracy of pose determination, while the larger number of outlier predictions with errors larger than $20$cm hamper the robustness of relocalization.}
\label{fig:CDF}
\end{figure}

The receptive field, denoted as $R$,  is an essential factor of Convolutional Neural Network (CNN) design. 
In our case, it determines how many image observations around a pixel are exposed and used for scene coordinate prediction.
Here, we would like to evaluate the impact of $R$ on the performance of SCoordNet.
SCoordNet presented in the main paper has $R=93$. We change the kernel size of  $7$-th to $12$-th layers of SCoordNet to adjust the receptive field to $29$, $45$, $61$, $125$, $157$, $189$, $221$, as shown in Table~\ref{table:params}.
Due to the time limitations, the evaluation only runs on \textit{heads} of \textit{7scenes} dataset \cite{shotton2013scene}. 
As reported in Table~\ref{table:receptive_field_result}, the mean of scene coordinate errors grows up as the receptive field $R$ decreases.
We illustrate the CDF of scene coordinate errors in Fig.~\ref{fig:CDF}.
It is noteworthy that a smaller $R$ results in more outlier predictions which cause a larger mean of scene coordinate errors.
However, a larger mean of scene coordinate error does not necessarily lead to a decrease in relocalization accuracy.
For example, a receptive field of $61$ has worse mapping accuracy than the larger receptive fields, but it achieves the smaller pose error and the better pose accuracy than them.
As we can see from Fig.~\ref{fig:CDF}, a smaller receptive field has a larger portion of precise scene coordinate predictions, especially those with errors smaller than $2cm$. 
These predictions are crucial to the accuracy of pose determination, as the outlier predictions are generally filtered by RANSAC.
Nevertheless, when we further reduce $R$ from $61$ to $45$ and then $29$, a drop of relocalization accuracy is observed. 
It is because, as $R$ decreases, the growing number of outlier predictions deteriorates the robustness of pose computation.
A receptive field between $45$ and $93$ is a good choice that respects the trade-off between precision and robustness.

\section{Ablation Study on the Downsample Rate} \label{sec:downsample_rate}

Due to the cost of dense predictions over full-resolution images, we predict scene coordinates for the images downsized by a factor of $8$ in the main paper, following previous works \cite{brachmann2018learning}. 
In this section, we intend to explore how the downsample rate affects the trade-off between accuracy and efficiency over SCoordNet. 
As reported in Table~\ref{table:params}, we change the kernel size and strides of $7$-th to $12$-th layers to adjust the downsample rate to $4$, $8$, $16$ and $32$ with the same receptive field of $93$.
The mean accuracy and the average time taken to localize frames of \textit{heads} are reported in Table~\ref{table:downsample}. 
As intuitively expected, the larger downsample rate generally leads to a drop of relocalization and mapping accuracy, as well as an increasing speed.
For example, the downsample rate $4$ and $8$ have a comparable performance, while the downsample rate $8$ outperforms $16$ by a large margin.
However, on the upside, a larger downsample rate is appealing due to the higher efficiency which scales quadratically with the downsample rate.
For real-time applications, a downsample rate of $32$ allows for a low latency of $34$ms per frame with a frequency of about $30$ Hz\footnote{All the experiments of this work run on a machine with a 8-core Intel i7-4770K, a 32GB memory and a
NVIDIA GTX 1080 Ti graphics card.}.

\begin{table}[t]
\centering
\resizebox{1\linewidth}{!}
{
\begin{tabular}{c|c|c|c|c|c}
\Xhline{2\arrayrulewidth}
\multirow{2}{*}{{\begin{tabular}[c]{@{}c@{}}Downsample\\ rate\end{tabular}}} & \multicolumn{2}{c|}{{Relocalization accuracy}} & \multicolumn{2}{c|}{{Mapping accuracy}} & \multirow{2}{*}{{Time}} \\ \cline{2-5}
                                                                                    & pose error           & pose accuracy         & mean              & stddev               &                                    \\ \hline \hline
4                                                                                   & \textbf{0.024m}, 0.97\textdegree          & \textbf{93.6\%}                          & \textbf{11.2cm} & 17.3                 & 1.34s                             \\ \hline
8                                                                                   & 0.024m, \textbf{0.91\textdegree}         & 92.9\%                           & 11.5cm                    & \textbf{16.4}                 & 0.20s                               \\ \hline
16                                                                                  & 0.025m, 0.92\textdegree         & 89.1\%                           & 16.3cm & 20.5                 & 0.11s                               \\ \hline
32                                                                                  & 0.029m, 1.06\textdegree         & 79.6\%                           & 20.7cm                    & 20.7                 & \textbf{0.034s}                              \\ 
\Xhline{2\arrayrulewidth}
\end{tabular}
}
\caption{The performance of SCoordNet \wrt the downsample rate. The pose accuracy means the percentage of poses with rotation and translation errors less than $5$\textdegree and $5$cm, respectively.}
\label{table:downsample}
\end{table}

\section{Running Time of KFNet Subsystems} \label{sec:time}
Table~\ref{table:timing_subsystems} reports the mean running time per frame (of size $640 \times 480$) of the measurement, process and filtering systems and NIS test, on a  NVIDIA GTX 1080 Ti. Since the measurement and process systems are independent and can run in parallel, the total time per frame is 157.18 ms, which means KFNet only causes an extra overhead of 0.58 ms compared to the one-shot SCoordNet.
Besides, our KFNet is 3 times faster than the state-of-the-art one-shot relocalization system DSAC++ \cite{brachmann2018learning}.

\begin{table}[h]
\resizebox{\linewidth}{!}
{
\begin{tabular}{c|c|c|c|c|c|c}
\Xhline{2\arrayrulewidth}
          & \multicolumn{5}{c|}{{KFNet}}                        & {DSAC++} \\ \hline \hline
Modules   & Measurement & Process & Filtering & NIS  & Total  & -      \\ \hline
Time (ms) & 156.60      & 51.23   & 0.29      & 0.29 & 157.18 & 486.07 \\
\Xhline{2\arrayrulewidth}
\end{tabular}
}
\caption{Running time of the subsystems of KFNet. }
\label{table:timing_subsystems}
\end{table}

\section{Mapping Visualization} \label{sec:mapping_visualization}

\begin{figure*}[]
\begin{center}
\includegraphics[width=0.85\linewidth]{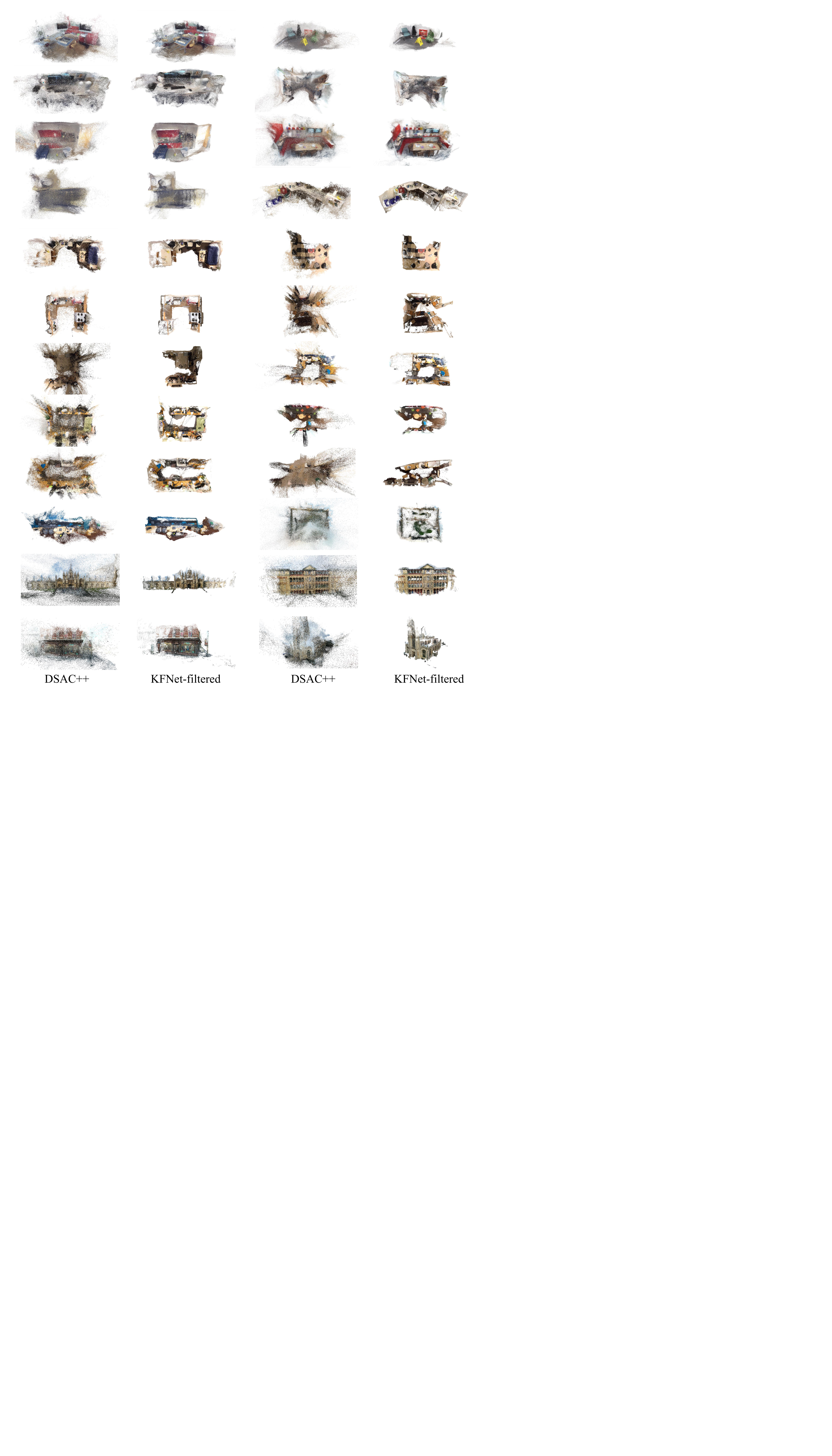}
\end{center}
  \caption{Point clouds of all the scenes predicted by DSAC++ \cite{brachmann2018learning} and our KFNet-filtered. Zoom in for better view.}
\label{fig:point_clouds_all}
\end{figure*}

As a supplement of Fig.~\ref{fig:point_clouds} in the main paper, we visualize the point clouds of \textit{7scenes} \cite{shotton2013scene}, \textit{12scenes} \cite{valentin2016learning} and \textit{Cambridge} \cite{kendall2015posenet} predicted by DSAC++ \cite{brachmann2018learning} and our KFNet-filtered in Fig.~\ref{fig:point_clouds_all}.
The clean point clouds predicted by KFNet in an end-to-end way provides an efficient alternative to costly 3D reconstruction from scratch \cite{zhu2017parallel,zhou2017progressive,zhang2017distributed,shen2018matchable,luo2018geodesc,zhu2018very,zhou2018learning,
luo2019contextdesc,zhang2019learning} in the relocalization setting, which is supposed to be valuable to mapping-based applications such as augmented reality.

\begin{table*}[p]
\centering
\resizebox{0.9\linewidth}{!}
{
\begin{tabular}{l|l|l|l}
\Xhline{2\arrayrulewidth}
{Input}  & {Layer}  & {Output} & {Output Size}              \\ \hline\hline
\multicolumn{4}{c}{\textbf{SCoordNet}} \\ \hline
$\mathbf{I}_t$  & Conv+ReLU, K=3x3, S=1, F=64   & conv1a & $\mathrm{H \times W \times 64}$  \\ \hline
conv1a & Conv+ReLU, K=3x3, S=1, F=64   & conv1b & $\mathrm{H \times W \times 64}$       \\ \hline
conv1b & Conv+ReLU, K=3x3, S=2, F=256  & conv2a & $\mathrm{H/2 \times W/2 \times 256}$  \\ \hline
conv2a & Conv+ReLU, K=3x3, S=1, F=256  & conv2b & $\mathrm{H/2 \times W/2 \times 256}$  \\ \hline
conv2b & Conv+ReLU, K=3x3, S=2, F=512  & conv3a & $\mathrm{H/4 \times W/4 \times 512}$  \\ \hline
conv3a & Conv+ReLU, K=3x3, S=1, F=512  & conv3b & $\mathrm{H/4 \times W/4 \times 512}$  \\ \hline
conv3b & Conv+ReLU, K=3x3, S=2, F=1024 & conv4a & $\mathrm{H/8 \times W/8 \times 1024}$ \\ \hline
conv4a & Conv+ReLU, K=3x3, S=1, F=1024 & conv4b & $\mathrm{H/8 \times W/8 \times 1024}$ \\ \hline
conv4b & Conv+ReLU, K=3x3, S=1, F=512  & conv5  & $\mathrm{H/8 \times W/8 \times 512}$  \\ \hline
conv5  & Conv+ReLU, K=3x3, S=1, F=256  & conv6  & $\mathrm{H/8 \times W/8 \times 256}$  \\ \hline
conv6  & Conv+ReLU, K=1x1, S=1, F=128  & conv7  & $\mathrm{H/8 \times W/8 \times 128}$  \\ \hline
conv7  & Conv, K=1x1, S=1, F=3    &   $\mathbf{z}_t $    & $\mathrm{H/8 \times W/8 \times 3}$ \\ \hline
conv7  & Conv+Exp, K=1x1, S=1, F=1    &   $\mathbf{V}_t$    & $\mathrm{H/8 \times W/8 \times 1}$    \\ \hline \hline
\multicolumn{4}{c}{\textbf{OFlowNet}} \\ \hline
$\mathbf{I}_{t-1} \|_0 \mathbf{I}_t$ & Conv+ReLU, K=3x3, S=1, F=16  & feat1 & $\mathrm{2 \times H \times W \times 16}$ \\ \hline
feat1 & Conv+ReLU, K=3x3, S=2, F=32  & feat2 & $\mathrm{2 \times H/2 \times W/2 \times 32}$ \\ \hline
feat2 & Conv+ReLU, K=3x3, S=1, F=32  & feat3 & $\mathrm{2 \times H/2 \times W/2 \times 32}$ \\ \hline
feat3 & Conv+ReLU, K=3x3, S=2, F=64  & feat4 & $\mathrm{2 \times H/4 \times W/4 \times 64}$ \\ \hline
feat4 & Conv+ReLU, K=3x3, S=1, F=64  & feat5 & $\mathrm{2 \times H/4 \times W/4 \times 64}$ \\ \hline
feat5 & Conv+ReLU, K=3x3, S=2, F=128  & feat6 & $\mathrm{2 \times H/8 \times W/8 \times 128}$ \\ \hline
feat6 & Conv, K=3x3, S=1, F=32  & $\mathbf{F}_{t-1} \|_0 \mathbf{F}_t$ & $\mathrm{2 \times H/8 \times W/8 \times 32}$ \\ \hline
$\mathbf{F}_{t-1} \|_0 \mathbf{F}_t$   & Cost Volume Constructor  & vol1 &  $\mathrm{H/8 \times W/8 \times w \times w \times 32}$ \\ \hline
vol1 &  Reshape & vol2 & $\mathrm{N \times w \times w \times 32 \;\; (N=HW/64)}$ \\ \hline
vol2 & Conv+ReLU, K=3x3, S=1, F=32  & vol3 & $\mathrm{N \times w \times w \times 32}$  \\ \hline
vol3 & Conv+ReLU, K=3x3, S=2, F=32  & vol4 & $\mathrm{N \times w/2 \times w/2 \times 32}$  \\ \hline
vol4 & Conv+ReLU, K=3x3, S=1, F=32  & vol5 & $\mathrm{N \times w/2 \times w/2 \times 32}$  \\ \hline
vol5 & Conv+ReLU, K=3x3, S=2, F=64  & vol6 & $\mathrm{N \times w/4 \times w/4 \times 64}$  \\ \hline
vol6 & Conv+ReLU, K=3x3, S=1, F=64  & vol7 & $\mathrm{N \times w/4 \times w/4 \times 64}$  \\ \hline
vol7 & Conv+ReLU, K=3x3, S=2, F=128  & vol8 & $\mathrm{N \times w/8 \times w/8 \times 128}$  \\ \hline
vol8 & Conv+ReLU, K=3x3, S=1, F=128  & vol9 & $\mathrm{N \times w/8 \times w/8 \times 128}$  \\ \hline
vol9 & Deconv+ReLU, K=3x3, S=2, F=64 & vol10 &  $\mathrm{N \times w/4 \times w/4 \times 64}$ \\ \hline
vol10 $\|_3$ vol7 & Conv+ReLU, K=3x3, S=1, F=64  & vol11 & $\mathrm{N \times w/4 \times w/4 \times 64}$ \\ \hline
vol11 & Deconv+ReLU, K=3x3, S=2, F=32 & vol12 &  $\mathrm{N \times w/2 \times w/2 \times 32}$ \\ \hline
vol12 $\|_3$ vol5 & Conv+ReLU, K=3x3, S=1, F=32  & vol13 & $\mathrm{N \times w/2 \times w/2 \times 32}$ \\ \hline
vol13 & Deconv+ReLU, K=3x3, S=2, F=16 & vol14 &  $\mathrm{N \times w \times w \times 16}$ \\ \hline
vol14 $\|_3$ vol3 & Conv+ReLU, K=3x3, S=1, F=16  & vol15 & $\mathrm{N \times w \times w \times 16}$ \\ \hline
vol15 & Conv, K=3x3, S=1, F=1 & confidence &  $\mathrm{N \times w \times w \times 1}$ \\ \hline
confidence & Spatial Softmax \cite{finn2015learning} & flow1 & $\mathrm{N \times 2}$ \\ \hline
flow1 & Reshape & flow2 & $\mathrm{H/8 \times W/8 \times 2}$ \\ \hline
flow2, $\hat{\boldsymbol{\theta}}_{t-1} \|_3 \boldsymbol{\Sigma}_{t-1}$ & Flow-guided Warping \cite{Zhu_2017_ICCV,Zhu_2018_CVPR,nguyen2018weakly,pfister2015flowing} & $\hat{\boldsymbol{\theta}}_t^- \|_3 \boldsymbol{\Sigma}_t^-$  & $\mathrm{H/8 \times W/8 \times 4}$  \\  
\Xhline{2\arrayrulewidth}
vol9 & Reshape & fc1 & $\mathrm{N \times 2w^2}$ \\  \hline
fc1 & FC+ReLU, F=64 & fc2 & $\mathrm{N \times 64}$ \\  \hline
fc2 & FC+ReLU, F=32 & fc3 & $\mathrm{N \times 32}$ \\  \hline
fc3 & FC+Exp, F=1 & fc4 & $\mathrm{N \times 1}$ \\  \hline
fc4 & Reshape & $\mathbf{W}_t$ & $\mathrm{H/8 \times W/8 \times 1}$ \\  \hline
\Xhline{2\arrayrulewidth}
\end{tabular}
}
\caption{ The full architecture of the proposed SCoordNet and OFlowNet. ``$\|_i$'' denotes concatenation along $i$-th dimension.}
\label{table:network}
\end{table*}

{\small
\bibliographystyle{ieee_fullname}
\bibliography{egbib}
}

\end{document}